\documentclass{article}

\usepackage{microtype}
\usepackage{graphicx}
\usepackage{subfigure}
\usepackage{booktabs} % for professional tables

\usepackage{hyperref}

\usepackage[accepted]{icml2024}

\usepackage{amsmath}
\usepackage{amssymb}
\usepackage{mathtools}
\usepackage{amsthm}

\usepackage{bm,bbm}
\usepackage{adjustbox}
\usepackage{multirow}
\usepackage{subfigure}
\usepackage{algorithm}
\usepackage{algorithmic}

\usepackage[capitalize,noabbrev]{cleveref}

\theoremstyle{plain}

\theoremstyle{definition}

\theoremstyle{remark}

\newcommand{\ourmodel}{\textsc{DeDer}}  % decompose and distillation of embodied reasoning capabilities

\newcommand{\StateSet}{\mathcal{S}}

\newcommand{\Goal}{\mathcal{G}}
\newcommand{\taskdesc}{\mathcal{H}}
\newcommand{\ActionSet}{\mathcal{A}}
\newcommand{\Dynamics}{P}

\DeclareMathOperator*{\argmax}{argmax}

\DeclareMathOperator*{\expectation}{\mathbb{E}}
\newcommand{\Dataset}{\mathcal{D}_{\text{exp}}}
\newcommand{\CoTDataset}{\mathcal{D}_{\text{Rtn}}}
\newcommand{\Rationale}{\mathcal{R}}
\newcommand{\LLM}{\Phi_{\text{LLM}}}
\newcommand{\sLM}{\Phi_{\text{sLM}}}
\newcommand{\RPolicy}{\Phi_{\text{R}}}
\newcommand{\DPolicy}{\Phi_{\text{P}}}
\newcommand{\KG}{g}

\usepackage[textsize=tiny]{todonotes}

\icmltitlerunning{Embodied CoT Distillation From LLM To Off-the-shelf Agents}

\begin{document}

\twocolumn[
\icmltitle{Embodied CoT Distillation From LLM To Off-the-shelf Agents}



\icmlsetsymbol{equal}{*}

\begin{icmlauthorlist}
\icmlauthor{Wonje Choi}{yyy}
\icmlauthor{Woo Kyung Kim}{yyy}
\icmlauthor{Minjong Yoo}{yyy}
\icmlauthor{Honguk Woo}{yyy}
\end{icmlauthorlist}

\icmlaffiliation{yyy}{Department of Computer Science and Engineering, Sungkyunkwan University, Suwon, Republic of Korea}

\icmlcorrespondingauthor{Honguk Woo}{hwoo@skku.edu}

\icmlkeywords{Machine Learning, ICML}

\vskip 0.3in
]



\printAffiliationsAndNotice{}  % leave blank if no need to mention equal contribution

\begin{abstract}
We address the challenge of utilizing large language models (LLMs) for complex embodied tasks, in the environment where decision-making systems operate timely on capacity-limited, off-the-shelf devices.
We present $\ourmodel$, a framework for decomposing and distilling the embodied reasoning capabilities from LLMs to efficient, small language model (sLM)-based policies. In $\ourmodel$, the decision-making process of LLM-based strategies is restructured into a hierarchy with a reasoning-policy and planning-policy. The reasoning-policy is distilled from the data that is generated through the embodied in-context learning and self-verification of an LLM, so it can produce effective rationales. The planning-policy, guided by the rationales, can render optimized plans efficiently. In turn, $\ourmodel$ allows for adopting sLMs for both policies, deployed on off-the-shelf devices. 
Furthermore, to enhance the quality of intermediate rationales, specific to embodied tasks, we devise the embodied knowledge graph, and to generate multiple rationales timely through a single inference, we also use the contrastively prompted attention model.
Our experiments with the ALFRED benchmark demonstrate that $\ourmodel$ surpasses leading language planning and distillation approaches, indicating the applicability and efficiency of sLM-based embodied policies derived through $\ourmodel$.
\end{abstract}

\section{Introduction}
\label{sec:intro}

In embodied AI, significant advancements have been made in applying large language models (LLMs) to task planning. For example, SayCan~\cite{llmagent:saycan} combines LLMs' reasoning capabilities with a reinforcement learning (RL)-based affordance model to interpret task instructions and deduce executable robotic skills in the environment. Several works~\cite{llmagent:zsp, llmagent:wu2023embodied, llmagent:llmplanner, llmagent:progprompt} explore the grounding of LLMs to the environment through prompting based on sensory data, reference trajectories, and available skills. Recently, palm-e~\cite{llmagent:palme} expands the embodied reasoning abilities of LLMs to include multimodal data, such as visual observations of the environment.
Yet, these approaches, which directly rely on LLMs for continual short-term decision-making, often encounter practical limitations in real-world applications, particularly when decision-making agents are required to operate on capacity-constrained, off-the-shelf devices. The high computational requirements of LLMs pose a significant challenge in such scenarios.

The direct end-to-end distillation of an LLM into a more compact, resource-efficient model, while it appears straightforward, might not be effective for complex embodied tasks~\cite{rldist:collaborating}.
This challenge stems from the requirements of a deep understanding on embodied task features, which inherently demand the long-horizon multi-step reasoning along with the ability to adapt to time-varying environment contexts. An embodied agent frequently encounters new and unseen environment information through its interaction with the surroundings. This continual exposure to a diverse range of environment conditions adds layers of complexity and variability, which in turn complicates the distillation process.

Our work is focused on the distillation of LLM-based policies for embodied tasks into off-the-shelf agents that are only capable of operating small language models (sLMs).
We present $\ourmodel$, an innovative embodied distillation framework, designed to decompose and distill the embodied reasoning and decision-making procedures of LLM-based policies into two distinct small, more manageable models: reasoning-policy and planning-policy. The reasoning-policy focuses on understanding and interpreting task requirements and environment contexts, while the planning-policy concentrates on generating actionable plans based on the insights provided by the reasoning-policy. This division allows for the sophisticated functionalities of LLMs to be leveraged in a more resource-efficient manner, suitable for embodied agents with capacity-limited, off-the-shelf devices.

Achieving the reasoning-policy via LLM distillation presents a unique challenge due to the hidden nature of reasoning processes within an LLM. We address this by employing the embodied Chain-of-Thought (CoT) and in-context learning, enhanced with self-verification, through the iterative use of the LLM.
For the reasoning-policy, we employ the embodied knowledge graph (KG)-based prompting and the contrastively prompted attention model, integrated with an sLM. These two techniques improve the quality of rationale outcomes from the reasoning-policy, by integrating the environment information to a KG efficiently and representing the current context effectively. 
They also allow for a parallel structure for multiple rationale generation, thereby facilitating the timely task planning at runtime.  
The planning-policy exploits the distilled rationales to determine executable plans, addressing the practical need for actionable decision-making for complex tasks. 

Using the ALFRED benchmark~\cite{ben:ALFRED20}, our experiments exhibit the advantages of $\ourmodel$. The results demonstrate that the policy derived through $\ourmodel$ significantly surpasses other baselines such as LLM-planner~\cite{llmagent:llmplanner} in zero-shot task planning scenarios. $\ourmodel$ achieves a substantial improvement of 15.0\% in seen task settings and 21.1\% in unseen settings. 
Considering that $\ourmodel$ employs an sLM at runtime instead of LLMs, the results clearly underline the exceptional adaptability of $\ourmodel$ in handling new and unencountered environments.

Note that $\ourmodel$ is the first framework to achieve sLM-based policies, which is resource-efficient yet comparable to LLM-based policies (i.e., the baselines in Section~\ref{subsec:exp}) in performance, for complex embodied tasks.
The contributions of our work are summarized as follows.
\begin{itemize}
    \item We present the novel framework $\ourmodel$, addressing the challenges of distilling LLMs' reasoning capabilities for embodied tasks to a small policy, readily deployed on capacity-limited, off-the-shelf devices. 
    \item We devise the two-tier policy hierarchy in  $\ourmodel$, through which the embodied reasoning process is decomposed and its knowledge can be distilled systematically to achieve a robust sLM-based policy. 
    \item We develop the data construction process from LLMs for rationales specific to embodied tasks, exploring in-context learning and self-verification techniques.  
    \item We implement the embodied KG and prompted attention model for sLM-based policies, to enhance the rationale quality across environment changes and facilitate rapid task planning.
    \item Through extensive experiments on ALFRED, we show  $\ourmodel$'s effectiveness and efficiency in achieving robust zero-shot performance for unseen embodied tasks.
\end{itemize}

\section{Related Work}
\label{sec:rel}

\noindent\textbf{LLM-based Embodied Control.}
In the field of embodied control, there is a growing trend of utilizing LLMs for reasoning and execution of tasks in real-world settings~\cite{llmagent:saycan, llmagent:zsp, llmagent:llmplanner}. Our work aligns with this direction but sets itself apart by aiming to enable off-the-shelf devices to attain comparable embodied task performance without directly utilizing LLMs at runtime, instead focusing on tuning a smaller language model.

\noindent\textbf{Embodied Policy Distillation.}
Recently, several works focused on distilling complex decision-making strategies, often derived from computationally intensive models, into compact and efficient ones suitable for resource-constrained environments.
In \cite{rldist:distilling}, knowledge is distilled from pre-trained vision-language models to supervise the language grounded skills of instruction-following agents. In \cite{rldist:gridtopix}, a two-stage training scheme is adopted for visual embodied agents. A relevant subset of policy distillation in RL is transferring the teacher policy in a supervised fashion~\cite{rldist:knowledge}. In particular, prior work concentrated on reducing the cross-entropy between the distributions of teacher and student policies~\cite{rldist:actor,rldist:kickstarting}. Our LLM-based policy distillation is also to minimize the divergence from the distribution of a teacher policy, which is an LLM, while exploring the unique two-tier hierarchy in decomposition and distillation of the LLM's reasoning capabilities.

\noindent\textbf{Reasoning Capabilities of LLMs.}
Numerous studies investigated the reasoning capabilities of LLMs, exploring methods like retrieval-augmented in-context examples~\cite{ra:rag, ra:incontext}, KG integration~\cite{ra:andrus2022enhanced, ra:baek2023knowledge}, and CoT prompting~\cite{cot:CoT, cot:selfconsist}. Recent research also demonstrated the effectiveness of distilling CoT processes from LLMs into sLMs~\cite{cot:scott, cot:symbolic}. Our work is in the same vein as LLM distillation, but specifically targets complex embodied tasks and uses decomposed distillation.

\begin{figure}[t]
% \vskip 0.2in
\begin{center}
   \includegraphics[width=\linewidth]{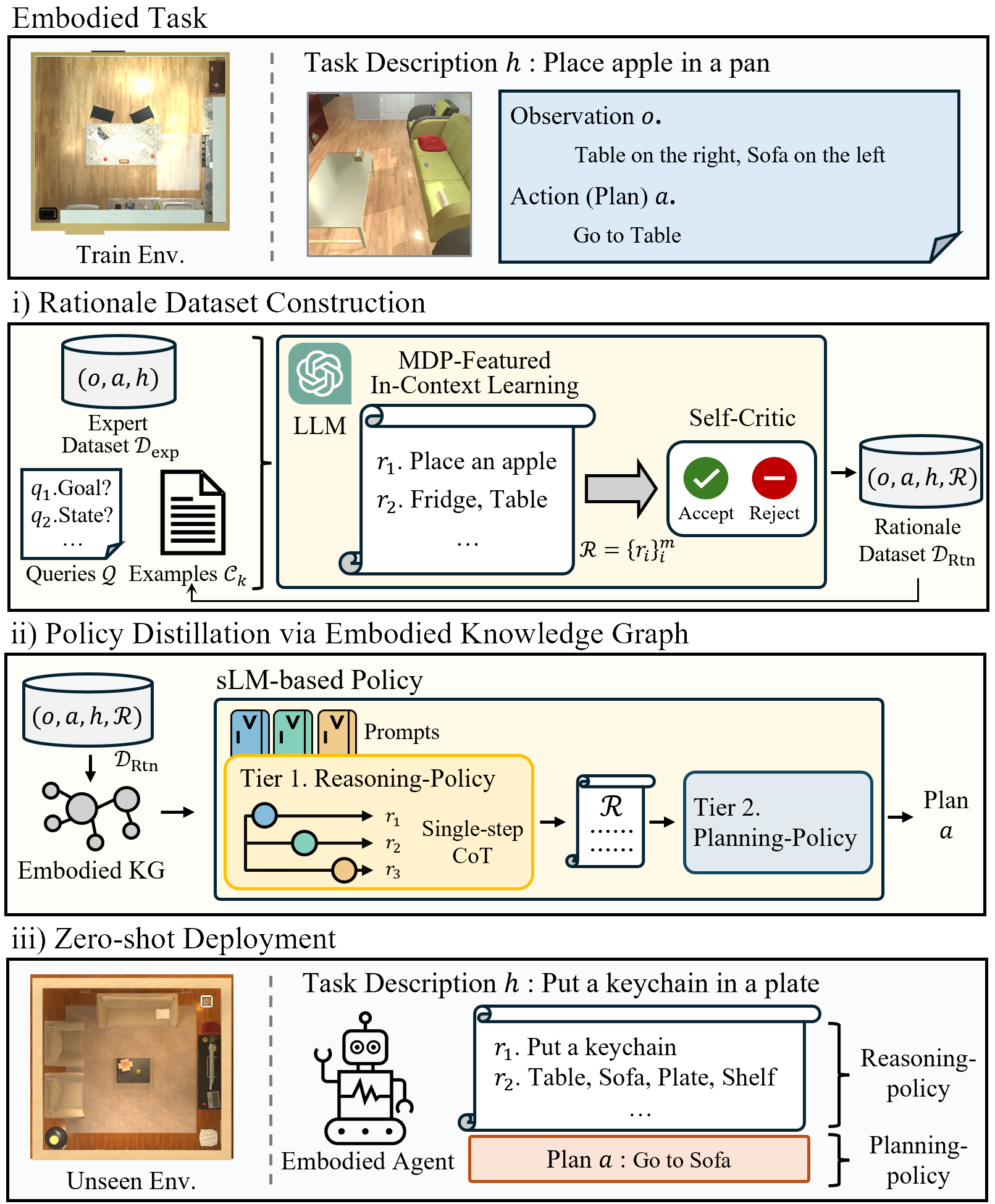}
   \vskip -0.05in
   \caption{$\ourmodel$ framework with three phases: (i) In rationale dataset construction phase, the MDP-featured in-context learning and self-critic function are employed to extract rationales from the LLM; (ii) In policy distillation phase, the sLM-based policy consisting of reasoning-policy and planning-policy is trained using the extracted rationale data; (iii) In zero-shot deployment, the distilled sLM-based policy is evaluated in unseen environments.}
   \label{fig:1}
\end{center}
\vskip -0.2in
\end{figure}

\section{Problem Formulation}
\label{sec:pre}
In RL, an environment for embodied agents is modeled as a Partially Observable Markov Decision Process (POMDP), represented by a tuple $(\StateSet, \ActionSet, \Dynamics, \Goal, \taskdesc, R, \Omega, \mathcal{O})$~\cite{llmagent:llmplanner, llmagent:progprompt}.
Here, $s \in \StateSet$ is a state space, $a \in \ActionSet$ is an action space, $\Dynamics: \StateSet \times \ActionSet \times \StateSet \rightarrow [0,1]$ is a transition probability, $G \in \Goal$ is a goal space, $h \in \taskdesc$ is a high-level task description and $R: \StateSet \times \ActionSet \times \Goal \rightarrow \mathbb{R}$ is a reward function.
The distinct aspect of embodied agents' environment lies in its nature of partial observations, featured as an observation space $o \in \Omega$ and a conditional observation probability $\mathcal{O}: \StateSet \times \ActionSet \rightarrow \Omega$~\cite{sutton2018reinforcement}.
This aspect accounts for the agents' limited perception, rendering the decision-making complex and reflective of real-world situations.
Our goal is to achieve a robust sLM-based policy $\sLM^*$ for capacity-limited, off-the-shelf devices, which is comparable to the capabilities in embodied task planning demonstrated by LLM-based policies $\Phi_\text{LLM}$.
\begin{equation}
\begin{aligned}
     \sLM^* = \argmax_{\sLM} & \expectation \Bigg[\sum_{t=0}^{\infty} \gamma^t R(s_t, \sLM(o_t, h_t), G) \\ 
     & - D(\LLM(o_t, h_t), \sLM(o_t, h_t) ) \Bigg]
\end{aligned}
\end{equation}
Note that $D$ is a distance function such as Kullback-Leibler divergence~\cite{kullback1951information} and $\gamma$ is a discount factor of the environment.

\section{Approach}
\label{sec:app}
For embodied tasks, it is essential for the agent to have reasoning capabilities to understand and interact with complex, dynamic environments. 
Yet, the simplification of the reasoning process is particularly necessary when employing an sLM-based policy, given the inherent limitations of sLMs due to their restricted model capacity. 
This can be achieved by integrating Markov Decision Process (MDP) features such as goal, state, observation, action, return-to-go, and sub-goal, which RL formulations specify, into the reasoning process~\cite{drl:gc,drl:pomdp,drl:dt,drl:tt}.

In this work, we refer to this type of environment information and MDP features as rationales, as they can function as justifications or hints that help to elaborate the reasoning behind plans.
We leverage these rationales as a means to effectively distill the embodied reasoning capabilities from an LLM to small models, thereby achieving an sLM-based policy. For this distillation, we develop the $\ourmodel$ framework comprising these phases: (\romannumeral 1) rationale dataset construction, (\romannumeral 2) policy distillation via embodied KG, and (\romannumeral 3) zero-shot deployment and evaluation, as illustrated in Figure~\ref{fig:1}.

In the phase of rationale dataset construction, we harness the CoT scheme inherent in the usage of LLMs to extract rationales from expert transitions (i.e., series of action plans) in the environment. 
This is achieved through MDP-featured in-context learning, employing RL-specific queries as prompts that are defined by the properties of the MDP.
In the subsequent phase of policy distillation, we establish an sLM-based policy structured in a two-tier hierarchy based on an embodied KG. It includes a reasoning-policy, which is trained to generate rationales in a single-step CoT optimized by behavior-based contrastive learning, as well as a planning-policy, which is learned to infer action plans through CoT prompting guided by these rationales.
In the deployment phase, we evaluate distilled sLM-policy in a zero-shot manner for unseen environments in which task descriptions, object positions, and indoor scenes are changed.

\begin{figure}[t]
% \vskip 0.2in
\centering
   \includegraphics[width=\linewidth]{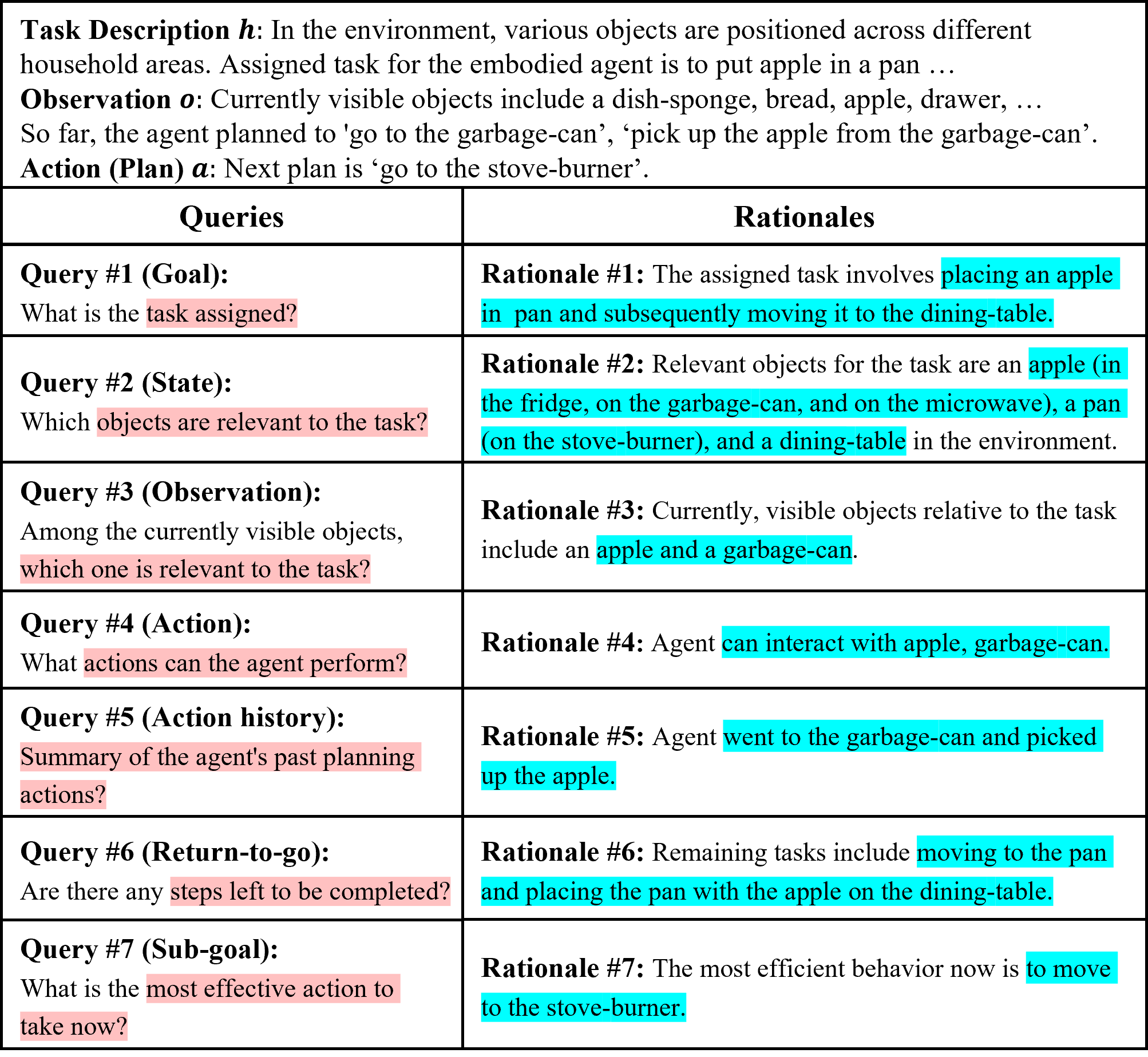}
   \vskip -0.05in
   \caption{MDP-featured in-context learning in $\ourmodel$ for rationale extraction from the LLM: the examples of inputs, queries (in red), and rationales (in blue) for the desired plan are presented, wherein MDP-aligned ones are specifically emphasized.}
   \label{fig:examples}
\vskip -0.2in
\end{figure}

\subsection{Rationale Dataset Construction} \label{subsec:datagen}
Consider an expert dataset $\Dataset=\{\tau_i=(o_i,a_i,h_i)\}_{i}$,
where each transition $\tau_i$ includes an observation $o_i$, action (plan) $a_i$, and high-level task description $h_i$ for timesteps $i$.
We expand the $\Dataset$ dataset to establish a rationale dataset $\CoTDataset=\{c_i=(o_i,a_i,h_i,\Rationale_i)\}_{i}$, where each transition $\tau_i$ is supplemented with a rationale set $\Rationale=\{r_j\}_{j=1}^{m}$.
To obtain the rationale set specifically configured for given embodied tasks, we integrate MDP-featured in-context learning with the CoT prompting mechanism of an LLM.
This involves iteratively prompting the LLM with a series of RL-specific queries, exploiting retrieval-augmented examples, similar to~\cite{ra:incontext}.
Subsequently, the rationale set undergoes LLM's assessments, as discussed in~\cite{cot:iter}, to be incorporated into the dataset $\CoTDataset$. 

\noindent\textbf{MDP-Featured In-Context Learning.}
To extract the rationales from the LLM using the transition $\tau$, we continually update in-context examples in a retrieval-augmented manner from dataset $\mathcal{D}_\text{Rtn}$. 
We use a retriever function $F:(\tau, \mathcal{C}) \mapsto \mathcal{C}_{k}$, as described in~\cite{ra:dpr}. It takes a transition $\tau$ from $\Dataset$ and a set of tuples $\mathcal{C}=\{c_1,...,c_n\}$ from $\CoTDataset$ as input, and retrieves the top-$k$ most semantically relevant tuples from $\mathcal{C}$ for given $\tau$, thus achieving an example set $\mathcal{C}_k$.
The semantic relevance is calculated by the inner product between language embeddings of $\tau$ and $c$ through the pre-trained 
contextual embedding model $E$. That is, we obtain relevance $S(\tau,c) = E(\tau)^{\top}E(c)$.

With the tuples $\mathcal{C}_k$, we then have the rationale set $\Rationale$ sequentially by prompting the LLM $\Phi_\text{LLM}$ with a pre-defined set of RL-specific queries $\mathcal{Q} = \{q_1, ..., q_m\}$.  
\begin{equation} \label{eq:1}
    \Rationale = \{r_l | r_l = \Phi_\text{LLM}(\mathcal{C}_{k}, \tau, \{r_j\}_{j < l}, \ q_l)\}
\end{equation}
Here, $\{r_j\}_{j<l}$ denotes a set of previously generated rationales for the questions preceding $r_{l}$. 
In this process, $\mathcal{C}_k$ is used to enhance the in-context learning of the LLM, as described in~\cite{ra:incontext}, enabling it to effectively respond to queries $q_l$.
In specific, RL-specific queries are designed to extract MDP features, which are necessary for embodied task planning such as a goal, state, plan, observation, plan history, and sub-goal.
The example of these queries and rationales is shown in Figure~\ref{fig:examples}.

\begin{figure*}[t]
\centering
   \includegraphics[width=0.90\linewidth]{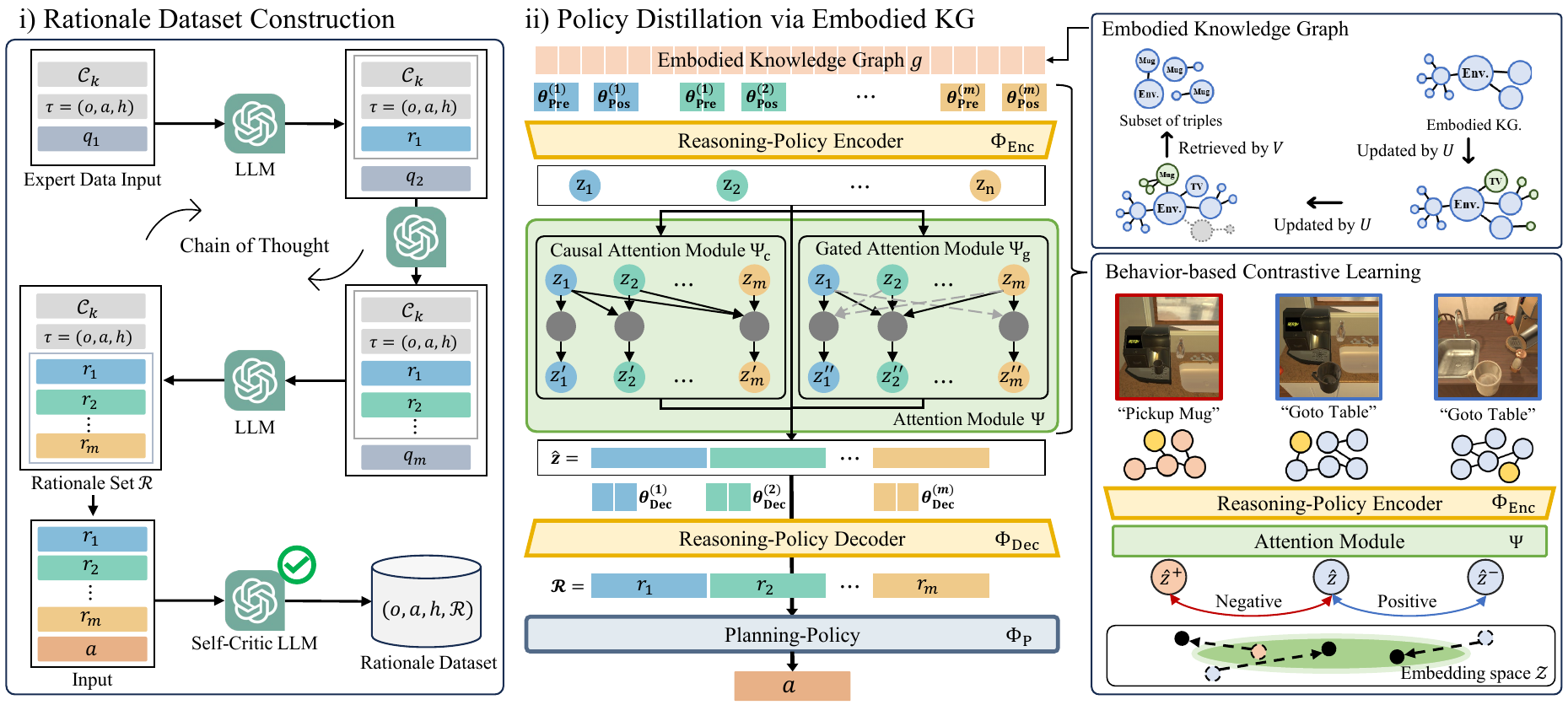}
   \vskip -0.05in
   \caption{Distillation procedures in $\ourmodel$:  
   During the rationale dataset construction phase, the LLM is iteratively prompted with queries $q_i$ and rationales $r_i$ to refine in-context examples $\mathcal{C}_k$ through retrieval augmentation. The LLM also serves as a critic, evaluating the validity of the extracted rationales $\mathcal{R}$. 
   %\color{red}
   During the policy distillation phase, the embodied KG containing environment information as well as expert experiences is used as input $g$ to the sLM-based reasoning-policy with the prompted casual attention, which is trained through behavior-based contrastive learning. The structure of reasoning-policy is specifically designed to produce multiple rationales $\mathcal{R}$ in a single-step CoT process through the integration of the prompted attention $\Psi$ and the encoder-decoder architecture. The reasoning-policy is distilled from the embodied KG, which is continually updated from the dataset. Subsequently, the planning-policy $\Phi_\text{p}$ is trained to produce a timely action plan $a$, by immediately using the rationales $\mathcal{R}$ at each step.  
   %our approach utilizes prompting from an embodied knowledge graph and supports single-step CoT with a prompted attention module. The reasoning-policy is optimized through behavior-based contrastive learning.
   %\color{black}
   }
   \label{fig:onecol}
\vskip -0.2in
\end{figure*}

\noindent\textbf{LLM as a Self-Critic Function.} 
To ensure that the rationale set $\mathcal{R}$ aligns with the action plan $a$, we also use the LLM as a self-critic function. Specifically, we use a query $q_{\text{cri}}$ to prompt the LLM to check whether the plan $a$ can be induced solely from the extracted $\mathcal{R}$. In cases when $\mathcal{R}$ does not provide sufficient information for $a$, we start over by retrieving in-context examples. Otherwise, we incorporate the newly generated tuple $c = (o, a, h, \mathcal{R})$ to the dataset $\CoTDataset$.
By employing this self-verification, we aim to gather rationales that encompass sufficient information to deduce plans in the expert transitions.
\begin{equation} \label{eq:2}
    \CoTDataset = \{c_i | \LLM(q_{\text{cri}}, \Rationale_i, a_i) = 1, c_i \in \CoTDataset \}
\end{equation}

\subsection{Policy Distillation via Embodied Knowledge Graph}
\label{subsec:pod}
To distill the reasoning capabilities of the LLM to an sLM-based policy $\sLM$ using the rationale dataset $\CoTDataset$, we structure the policy in a two-tier hierarchy. The first tier is a reasoning-policy $\RPolicy$; it is responsible for inferring a rationale set from a given observation $o$, a task description $h$, and an embodied KG $\KG$.
The second tier is a planning-policy $\DPolicy$; it generates the plan, guided by the rationales from $\RPolicy$.
\begin{equation}
    \sLM = \DPolicy \circ \RPolicy : (o, h; \KG) \mapsto a.
\end{equation}
The embodied KG is an internal component of the sLM-based policy, encapsulating the environment information.

In training, we use a fine-tuning method with soft prompts for the sLM-based policy. This is effective for adopting sLMs with limited reasoning capabilities, where in-context learning is not allowed.

\noindent\textbf{Embodied KG.}
As the agent continuously interacts with the environment and can accumulate information for task completion, it is important to represent information efficiently for prompting the sLM-based policy.
We employ an embodied KG, a set of triples $\KG=\{x_i=(x^s_i,x^r_i,x^o_i)\}_i$, where $x^s$ is the subject, $x^r$ is the relation, and $x^o$ is the object. For instance, given ``an apple is on the table'' and ``the agent picks up a knife'', the corresponding triples are (\textit{Apple}, \textit{On}, \textit{Table}) and (\textit{Agent}, \textit{Pickup}, \textit{Knife}), respectively.
We refine the embodied KG at each planning step $t$ by an update function $U$ such as 
\begin{equation}
\label{eq:update}
    U : (\KG_{t-1}, \ a_{t-1}, \ o_t) \mapsto \KG_t.
\end{equation}
To prompt the sLM-based policy, we also use the KG retriever function $V$, which retrieves a subset of triples from $\KG$ relevant to observation $o$ and task description $h$. 
\begin{equation}
\label{eq:kgretriever}
    V : (o,h;\KG) \rightarrow \{x\in \KG|S(x, (h, o)) \geq \delta\}
\end{equation}
The relevant triples are chosen by the pre-trained semantic relevance function $S$ between each triple in $\KG$ and inputs $o$ and $h$, where $\delta$ is a threshold hyperparameter. Hereafter, $g$ denotes the graph extracted via the KG retriever function.

\noindent\textbf{Reasoning-Policy Distillation.}
For the reasoning-policy $\Phi_{\text{R}}$ which produces a rationale set, we employ the attention module with an encoder-decoder architecture.
\begin{equation}
    \RPolicy = \Phi_{\text{Dec}} \circ \Psi \circ \Phi_{\text{Enc}} : \KG \mapsto \mathcal{R}
\end{equation}
To facilitate the single-step CoT through the reasoning-policy $\Phi_\text{R}$, we also use soft prompt pools $\theta = [\theta^{(1)}, \theta^{(2)}, ..., \theta^{(m)}],\ \theta^{(i)} \in \mathbb{R}^{d}$, where $d$ is the dimension of prompt $\theta^{(i)}$~\cite{t5prompt}.
The encoder $\Phi_{\text{Enc}}$ incorporates two distinct prompt pools: a prefix prompt pool $\theta_\text{Pre}$ and a postfix prompt pool $\theta_\text{Pos}$.
\begin{equation}
    \Phi_{\text{Enc}} : (\KG ; \theta_{\text{Pre}}, \theta_{\text{Pos}}) \mapsto z = [z_1, ..., z_d]
\end{equation}
Each prefix prompt $\theta_{\text{Pre}}^{(i)}$ is initialized based on the language embedding of the query $q_i$, while each postfix prompt $\theta_{\text{Pos}}^{(i)}$ is randomly initialized.
Furthermore, for emphasizing information in each rationale and transferring it sequentially, in line with the rationale dataset construction, the attention module $\Psi$ includes a causal attention  $\Psi_\text{c}$~\cite{attn:causal} and gated attention $\Psi_\text{g}$~\cite{attn:gated}, i.e., 
\begin{equation}
\hat{z} = [\hat{z}_1, ..., \hat{z}_d] = \Psi(z) = z + \alpha (\Psi_\text{c}(z) + \Psi_\text{g}(z))
\end{equation}
where $\alpha$ is a scaling factor that regulates the influence of the attention mechanisms' outputs.
The decoder $\Phi_{\text{Dec}}$ utilizes a decoder prompt pool  $\theta_\text{Dec}$ to generate a set of rationales $\mathcal{R}$. 
\begin{equation}
\Phi_{\text{Dec}} : (\hat{z} ; \theta_\text{Dec}) \mapsto \mathcal{R}
\end{equation}

With the embodied KG generated by update function $U$ and KG retriever function $V$ from $\mathcal{D}_\text{Rtn}$, we optimize the reasoning-policy by the rationale reconstruction loss.
% to optimize the generation of each rationale in the training dataset. 
\begin{equation}
\begin{aligned}
    \mathcal{L}_{\text{Rtn}}
     = \expectation_{(o,h,\mathcal{R}) \sim \CoTDataset} \left[ \sum_{i=1}^{m} \log \Phi_{\text{R}}(r_i|\KG) \right]
\label{loss:RationaleGenerationLoss}
\end{aligned}
\end{equation}
This loss is calculated as the expected sum of the log-likelihoods for generating each rationale $r_i$.

Considering that subtle changes in the environment might lead to inconsistent agent's plan, we devise the prompted KG representations, using behavior-based contrastive learning~\cite{drl:atc, drl:aco, drl:conpe}.
The prompted KG representations facilitate the causal and gated attentions for the reasoning-policy, thus enabling a single-step inference for multiple rationales.
We sample a batch of embodied KG pairs $\mathcal{B}_\text{Con} = \{(\KG_i, \KG_i^{+}), (\KG_i, \KG_i^{-})\}_{i}$, where $(\KG_i, \KG_i^{+})$ denotes a positive pair, and $(\KG_i, \KG_i^{-})$ denotes a negative pair.
% and $I: (\KG, \KG) \mapsto \{0, 1\}$ is contrast function.
Specifically, the positive pair consists of embodied KG executing the same plan, while the negative pair is defined as consecutive planning steps.
Then, the contrastive learning loss is formulated as
\begin{equation}
% \begin{aligned}
    \mathcal{L}_{\text{Con}}=  
    \expectation_{\mathcal{B}_\text{Con} \sim \mathcal{D}_\text{Rtn}} [\text{max}\{0, d(\hat{z}, \ \hat{z}^{+})
     - d(\hat{z},\ \hat{z}^{-}) + \epsilon\}]
% \end{aligned}
\label{loss:ContrastiveLoss}
\end{equation}
where $\hat{z} = \Psi \circ \Phi_\text{Enc}(\KG;\theta_{\text{Pre}}$, $\theta_{\text{Pos}})$, $d$ represents the sum of a distance metric within the embedding space $\hat{z} \in \mathcal{Z}$  corresponding to an element of the rationale embedding sequence~\cite{con:simclr, con:infonce}, and $\epsilon$ is a margin parameter. 

\noindent\textbf{Planning-Policy Distillation.}
\label{subsec:}
The planning-policy $\Phi_{\text{P}}$ predicts a next plan $a$ based on the rationale set generated from the reasoning-policy $\RPolicy$.
\begin{equation}
    \DPolicy : (\mathcal{R} = \RPolicy(\KG)) \mapsto a
\end{equation}
We optimize the planning-policy via the reconstruction loss.
\begin{equation}
    \mathcal{L}_{\text{Plan}} =  
    \expectation_{(o,a,h) \sim \CoTDataset, \mathcal{R} \sim \RPolicy} \left[ \log \DPolicy(a \ | \ \mathcal{R}) \right]
\label{loss:PlanGenerationLoss}
\end{equation}
Algorithm~\ref{alg:distil&plan} lists the policy distillation procedures, where the losses in~\eqref{loss:RationaleGenerationLoss}, \eqref{loss:ContrastiveLoss} and the loss in~\eqref{loss:PlanGenerationLoss} are used for the reasoning policy and the planning-policy, respectively.

\begin{algorithm}[t]
\caption{Policy Distillation}
\label{alg:distil&plan}
Rationale Dataset $\CoTDataset$ \\
Initialize reasoning-policy $\RPolicy$, and planning-policy $\DPolicy$ \\
Initialize prompt pools $\theta_\text{Pre}, \theta_\text{Pos}, \theta_\text{Dec}$
\begin{algorithmic}[1]
% phase1
\STATE \textit{/* Reasoning-Policy Distillation */}

\WHILE{not converge}
    % KG
    % \STATE{Sample a batch $\mathcal{B} \sim \CoTDataset$}
    \STATE{Sample a batch $\mathcal{B} = \{(o_i,a_i,h_i)\}_{i} \sim \CoTDataset$}
    \STATE{Obtain $\mathcal{B}_\text{Con} = \{(\KG_i, \KG_i^{+}), (\KG_i, \KG_i^{-})\}_{i}$ using \eqref{eq:update}, \eqref{eq:kgretriever}}
    % \STATE{Calculate loss $\mathcal{L}_{\text{R}}$ using~\eqref{loss:RationaleGenerationLoss}}
    \STATE{Update $\RPolicy, \theta_\text{Pre}, \theta_\text{Pos}, \theta_\text{Dec}$ using \eqref{loss:RationaleGenerationLoss}, \eqref{loss:ContrastiveLoss}}
\ENDWHILE
% \FOR{each $c$ in $\CoTDataset$}
%     \STATE{Predict rationales $\mathcal{R} \leftarrow \Phi_\text{R}(o, h)$}
%     \STATE{Update CoT sample $c \leftarrow (o,a,h,\mathcal{R})$}
% \ENDFOR
% phase2
\STATE \textit{/* Planning-Policy Distillation */} \\
\WHILE{not converge}
    \STATE{Sample a batch $\mathcal{B} = \{(o_i,a_i,h_i)\}_i \sim \CoTDataset$}
    \STATE{Obtain $\mathcal{B} = \{(a_i, g_i)\}_{i}$ using \eqref{eq:update}, \eqref{eq:kgretriever}}
    \STATE{Calculate $\Rationale$ using $\RPolicy$ on batch $\mathcal{B}$}
    \STATE{Update $\DPolicy$ using~\eqref{loss:PlanGenerationLoss} }
\ENDWHILE
\end{algorithmic}
\end{algorithm}
\color{black}

\begin{table*}[h]
\caption{Performance of embodied task planning in ALFRED with $4$ different task categories}
\vskip 0.1in
\begin{center}
\begin{adjustbox}{width=0.95\textwidth}
    \begin{tabular}{l cc cc cc cc}
    \toprule
    \multirow{2}{*}{Method}
    & \multicolumn{2}{c}{Train} & \multicolumn{2}{c}{Seen} & \multicolumn{2}{c}{Unseen Spatial} & \multicolumn{2}{c}{Unseen Environment} \\
    
    \cmidrule(rl){2-3} \cmidrule(rl){4-5} \cmidrule(rl){6-7} \cmidrule(rl){8-9}
     & SR  & GC  & SR  & GC  & SR  & GC  & SR  & GC  \\
    \midrule
    \multicolumn{8}{l}{\textbf{LLM-based policy}: PaLM (540B), LLaMA2 (7B)} & \\
    \midrule
    SayCan-PaLM%~\cite{llmagent:saycan}
    & $34.1{\pm0.0}$ & $68.8{\pm0.0}$
    & $37.6{\pm0.0}$ & $66.8{\pm0.0}$
    & $26.5{\pm0.0}$ & $66.5{\pm0.0}$
    & $29.3{\pm0.0}$ & $\mathbf{68.8{\pm0.0}}$ \\
    LLM-planner-PaLM %~\cite{llmagent:llmplanner}
    & $70.9{\pm0.0}$ & $86.6{\pm0.0}$
    & $66.8{\pm0.0}$ & $84.3{\pm0.0}$
    & $33.6{\pm0.0}$ & $67.6{\pm0.0}$
    & $17.2{\pm0.0}$ & $54.3{\pm0.0}$ \\
    ZSP-PaLM %~\cite{llmagent:zsp}
    & $73.8{\pm0.0}$ & $89.2{\pm0.0}$
    & $59.6{\pm0.0}$ & $80.7{\pm0.0}$
    & $28.8{\pm0.0}$ & $66.5{\pm0.0}$
    & $6.9{\pm0.0}$ & $36.5{\pm0.0}$ \\
    SayCan-LLaMA2
    & $0.0{\pm0.0}$ & $10.6{\pm0.0}$
    & $0.3{\pm0.0}$ & $9.6{\pm0.0}$
    & $0.0{\pm0.0}$ & $10.3{\pm0.0}$
    & $0.0{\pm0.0}$ & $2.2{\pm0.0}$ \\
    LLM-planner-LLaMA2
    & $1.8{\pm0.0}$ & $19.6{\pm0.0}$
    & $2.0{\pm0.0}$ & $22.7{\pm0.0}$
    & $0.8{\pm0.0}$ & $19.6{\pm0.0}$
    & $0.0{\pm0.0}$ & $15.8{\pm0.0}$ \\
    ZSP-LLaMA2
    & $54.3{\pm0.0}$ & $76.5{\pm0.0}$
    & $26.7{\pm0.0}$ & $59.9{\pm0.0}$
    & $6.7{\pm0.0}$ & $46.9{\pm0.0}$
    & $0.0{\pm0.0}$ & $26.6{\pm0.0}$ \\

    \midrule
    \multicolumn{8}{l}{\textbf{sLM-based policy}: GPT2-large (0.8B), GPT2 (0.2B)} & \\
    \midrule
    SayCan-GPT2-large
    & $0.2{\pm0.0}$ & $14.7{\pm0.0}$
    & $0.5{\pm0.0}$ & $17.1{\pm0.0}$
    & $0.5{\pm0.0}$ & $17.6{\pm0.0}$
    & $0.0{\pm0.0}$ & $18.1{\pm0.0}$ \\
    LLM-planner-GPT2-large
    & $0.0{\pm0.0}$ & $3.43{\pm0.0}$
    & $0.0{\pm0.0}$ & $4.0{\pm0.0}$
    & $0.0{\pm0.0}$ & $2.0{\pm0.0}$
    & $0.0{\pm0.0}$ & $1.8{\pm0.0}$ \\
    ZSP-GPT2-large
    & $1.8{\pm0.0}$ & $3.6{\pm0.0}$
    & $0.8{\pm0.0}$ & $3.4{\pm0.0}$
    & $0.3{\pm0.0}$ & $4.3{\pm0.0}$
    & $0.0{\pm0.0}$ & $0.4{\pm0.0}$ \\
    
    End2End-GPT2-large
    & $41.1{\pm12.2}$ & $63.5{\pm11.6}$
    & $25.2{\pm7.0}$ & $54.3{\pm10.9}$
    & $11.4{\pm4.5}$ & $50.1{\pm9.2}$
    & $5.7{\pm1.0}$ & $53.8{\pm25.3}$ \\

    SCoTD-GPT2%~\cite{cot:symbolic}
    & $55.8{\pm4.2}$ & $82.7{\pm1.5}$
    & $51.8{\pm5.2}$ & $79.0{\pm2.3}$
    & $29.3{\pm2.1}$ & $70.4{\pm0.9}$
    & $27.6{\pm1.7}$ & $59.8{\pm1.7}$ \\

    SCOTT-GPT2%~\cite{cot:scott}
    & $62.2{\pm1.6}$ & $85.6{\pm0.1}$
    & $57.2{\pm4.0}$ & $81.3{\pm1.6}$
    & $32.7{\pm2.1}$ & ${72.0\pm0.1}$
    & ${24.1\pm7.9}$ & $60.3{\pm1.6}$ \\
    
    End2End-GPT2
    & $33.1{\pm4.6}$ & $46.6{\pm8.1}$
    & $17.6{\pm2.6}$ & $38.8{\pm8.0}$
    & $8.5{\pm6.3}$ & $36.3{\pm9.1}$
    & $3.4{\pm0.9}$ & $34.6{\pm9.0}$ \\

    $\ourmodel$-GPT2
    & $\mathbf{100.0{\pm0.0}}$ & $\mathbf{100.0{\pm0.0}}$
    & $\mathbf{81.8{\pm0.5}}$ & $\mathbf{92.2{\pm0.2}}$
    & $\mathbf{52.7{\pm1.0}}$ & $\mathbf{81.2{\pm0.4}}$
    & $\mathbf{40.3{\pm0.9}}$ & $68.7{\pm0.6}$ \\

    \bottomrule
    \end{tabular}

\end{adjustbox}
\end{center}
\label{tab:main}
\vskip -0.1in
\end{table*}

\section{Evaluation}
\label{tab:exp}

\subsection{Experiment Setting}\label{subsec:exp}
\noindent\textbf{Environments.}
For evaluation, we use the ALFRED~\cite{ben:ALFRED20} environment.
For embodied reasoning tasks, ALFRED features a wide variety of interactive elements including $58$ distinct object types (e.g., bread) and $26$ receptacles object types (e.g., plate) across $120$ different indoor scenes (e.g., kitchen).
By combining these objects and indoor scenes with instructions of $7$ different types (e.g., pick \& place), $4703$ distinct tasks can be configured (e.g., ``Put a keychain in a plate and then put them in a shelf'').
This setup provides a broad spectrum of real-world-like challenges, encompassing complex navigation, object manipulation, and executing sequential operations.

We use $312$ trajectories for the expert dataset and organize the evaluation tasks into $4$ categories based on their similarities to the tasks in the expert dataset.
For \textbf{Train} category, the tasks are identical to those in the expert dataset. 
For \textbf{Seen} category, the tasks remain the same as those in the expert dataset, except that the starting positions of the task-irrelevant objects are placed randomly.
For \textbf{Unseen Spatial} category, all objects in the environment are placed randomly.
The most challenging category \textbf{Unseen Environment} includes new tasks and indoor scenes not presented in the expert dataset.
The environment details are in Appendix A.

\noindent\textbf{Baselines.} For comparison, we implement several language planning approaches:
1) \textbf{SayCan}~\cite{llmagent:saycan} is an embodied planning framework that integrates the probability from an LLM with affordance scores. For embodied control, the affordance is based on object presence information.
2) \textbf{ZSP}~\cite{llmagent:zsp} employs a step-wise planning to accomplish the embodied tasks.
3) \textbf{LLM-planner}~\cite{llmagent:llmplanner}, directly utilizes an LLM for embodied task planning, which dynamically re-plans when it fails to generate an executable plan.
In evaluating in off-the-shelf devices, we adopt sLMs for these language planning baselines (SayCan, ZSP, LLM-planner). 

We also implement several knowledge distillation algorithms:
4) \textbf{SCoTD}~\cite{cot:symbolic} is a knowledge distillation algorithm to train an sLM using reasoning samples derived from an LLM.
5) \textbf{SCOTT}~\cite{cot:scott} is a knowledge distillation method to train an sLM, which involves self-consistent CoT augmentation from an LLM and counterfactual reasoning objectives.
6) \textbf{End2End}~\cite{lmagent:butlers} is an embodied task planning method using a single-tier policy unlike $\ourmodel$, which directly generates a plan from the inputs.
To evaluate the task planning performance in the environment through generated trajectories, we also implement an additional rule-based policy that directly interacts with the environment, following the action plans from the baselines and our $\ourmodel$. 

\noindent\textbf{Evaluation metrics.} 
We use two different metrics in ALFRED~\cite{ben:ALFRED20}.
\textbf{Task Success Rate (SR)} (\%) is the percentage of tasks fully completed, where a task is regarded as a success if and only if all the sub-goals are achieved. For example, the task ``Slice a heated bread'' is decomposed into individual sub-goals like ``slice the bread'' and ``heat the bread''.
\textbf{Goal-conditioned Success Rate (GC)} (\%) is the percentage of sub-goals that are completed.

\subsection{Performance Evaluation}
In Table~\ref{tab:main}, we evaluate the embodied task planning performance, wherein each policy is evaluated in a zero-shot manner.
Our $\ourmodel$ consistently demonstrates the robust performance in both SR and GC metrics across all test categories (Train, Seen, Unseen Spatial, Unseen Environment), achieving $21.6\%$ higher SR and $12.3\%$ higher GC on average over the most competitive baseline LLM-planner-PaLM.
Given that LLM-planner-PaLM exploits the PaLM~\cite{palm} with $540$ billion parameters, $2700$ times larger than $\ourmodel$, this performance gain of $\ourmodel$ is particularly significant.
Moreover, compared to the baselines that have the same parameter size, we observe that $\ourmodel$ outperforms these baselines for all categories up to $27.6\%$ higher SR and $12.6\%$ higher GC on average.

The language planning baselines (SayCan, LLM-Planner, ZSP), which are configured to adopt sLMs (LLaMA2, GPT2-large), exhibit low performance. This is due to sLMs' limited reasoning capabilities.
Meanwhile, the knowledge distillation baselines (SCoTD, SCOTT) maintain decent performance. While they use distillation from an LLM via few-shot prompting, their distilled knowledge is somewhat limited by the conventional CoT mechanism. This limitation arises because they do not employ multi-step prompting and self-verification, unlike $\ourmodel$. 
Furthermore, the End2End baseline exhibits significantly low performance in directly conducting embodied task planning with the expert dataset, due to the limited reasoning capability of the sLM. 

In contrast, our framework employs the rationale dataset and the sLM-based policy with a two-tier hierarchy structure. This enables the effective distillation of the LLM's reasoning capabilities, specifically tailored for embodied task planning based on MDP-featured in-context learning. 

\subsection{Ablation Studies}
In the ablation studies, the performance metrics for all test categories (Train, Seen, Unseen) are reported in SR.

\noindent\textbf{Rationale Dataset Construction.}
For extracting rationales and constructing the dataset in Section~\ref{subsec:datagen}, we test several language models, including sLMs such as GPT2-large~\cite{gpt2} denoted as GPT2, and LLMs such as PaLM and GPT3~\cite{palm, gpt3}.
We also evaluate the dataset construction process without employing MDP-featured in-context learning and self-verification; This ablated method is denoted as \textit{Few-shot}, as described in~\cite{cot:CoT}, where a fixed set of examples is used for prompting rationale extraction.

In Table~\ref{tab:ab1}, there is a notable performance drop across the task categories when employing GPT2. These results are consistent with our motivation to harness the reasoning capabilities of LLMs for rationale extraction, which in turn contributes to the effective distillation into the sLM-based policy.
Moreover, $\ourmodel$ yields better performance compared to \textit{Few-shot} by an average of $5.35\%$ in the Unseen settings, excluding GPT2 results. This improvement indicates the benefits of our MDP-featured in-context learning and self-verification methods.

\begin{table}[H]
\caption{Ablation on rationale dataset construction}
\vskip 0.1in
\label{tab:ab1}
\centering
\begin{adjustbox}{width=0.95\linewidth}
\begin{tabular}{l c c c c}
    \toprule
    \small
    Method & LM & Train & Seen & Unseen \\
    \midrule

    Few-shot & GPT2
    & $41.9{\pm19.3}$
    & $2.4{\pm0.5}$
    & $0.3{\pm0.1}$ \\

    $\ourmodel$ & GPT2
    & $60.8{\pm4.0}$
    & $53.5{\pm2.2}$
    & $23.4{\pm8.2}$ \\
    \midrule

    Few-shot & GPT$3$
    & $100.0{\pm0.0}$
    & $72.8{\pm0.1}$
    & $38.6{\pm1.5}$  \\

    $\ourmodel$ & GPT$3$
    & $100.0{\pm0.0}$
    & $72.8{\pm0.1}$
    & $42.2{\pm1.2}$ \\

    Few-shot & PaLM
    & $100.0{\pm0.0}$
    & $76.9{\pm0.3}$
    & $39.6{\pm0.1}$ \\

    $\ourmodel$ & PaLM
    & $100.0{\pm0.0}$
    & $\mathbf{81.8{\pm0.5}}$
    & $\mathbf{46.5{\pm1.0}}$ \\
    
    \bottomrule
    \end{tabular}
\end{adjustbox}
\vskip -0.1in
\end{table}

\begin{figure}[h]
    % \vskip -0.1in
    \centering
    \subfigure[Rationale omission]{
        \centering
        \includegraphics[width=0.46\linewidth]{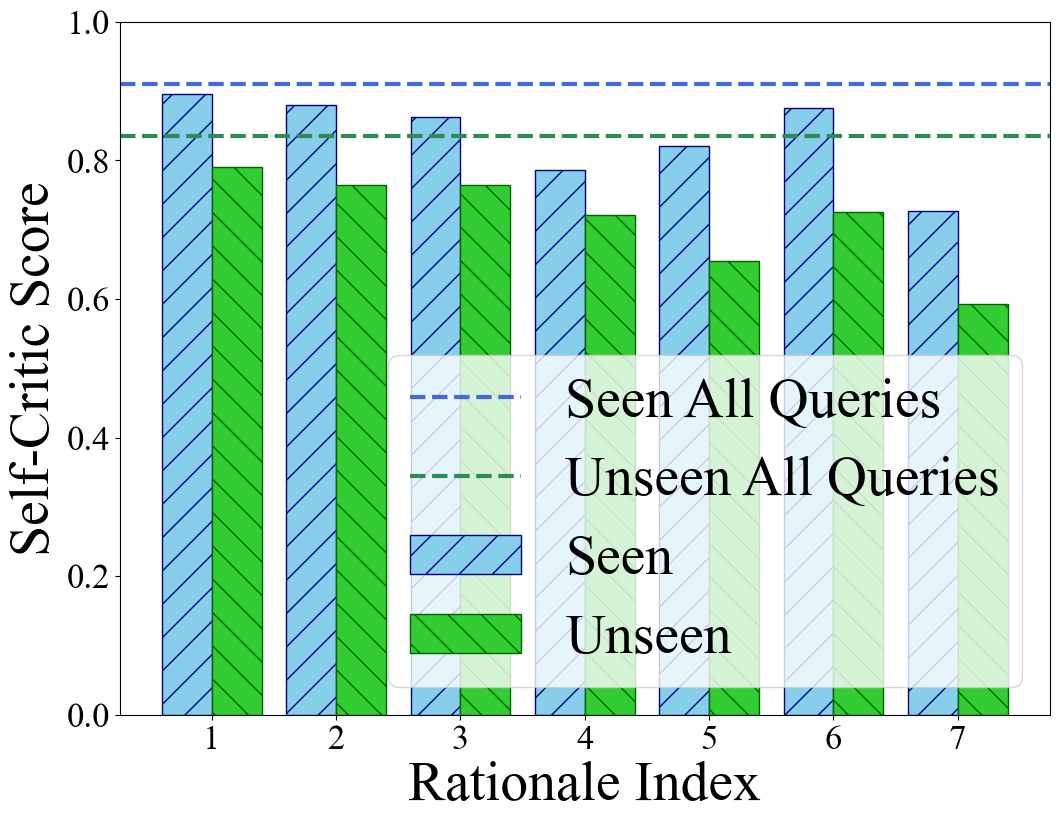}
        \label{fig:ab2_1}
    }
    % \hspace{10pt}
    \subfigure[Incomplete rationales]{
        \centering
        \includegraphics[width=0.46 \linewidth]{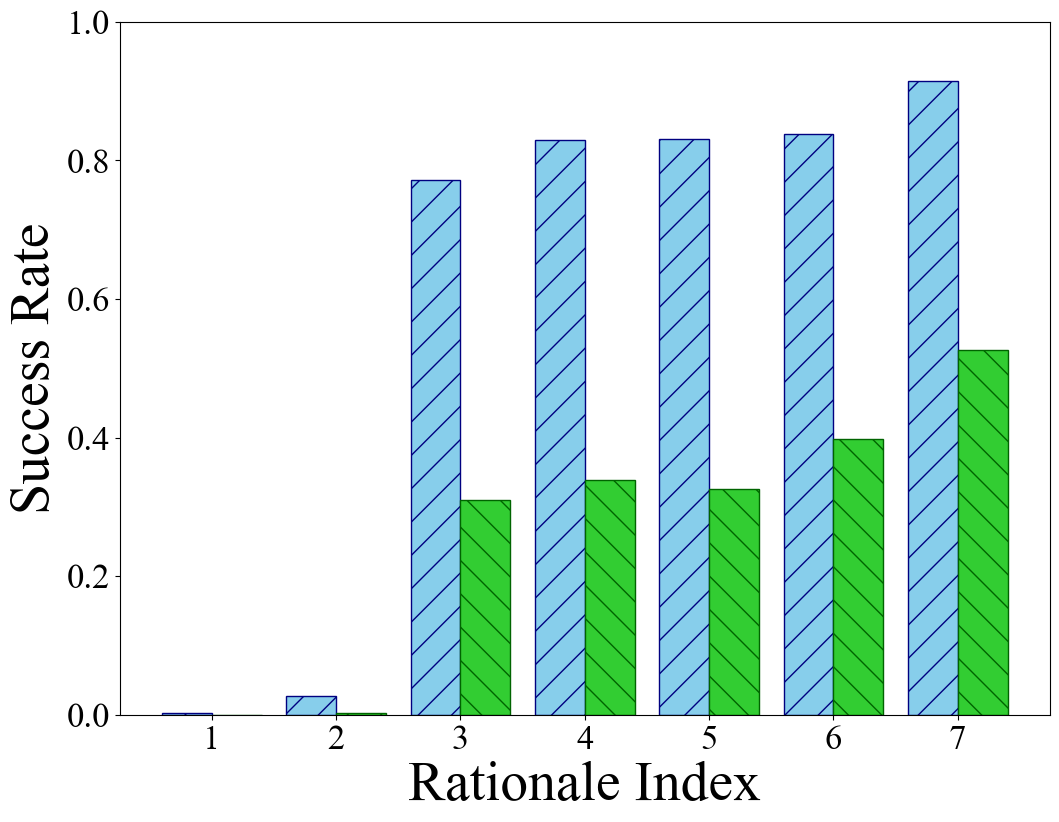}
        \label{fig:ab2_2}
    }
    \label{fig:ab2}
    \vskip -0.05in
    \caption{Ablation on each rationale}
    % \vskip -0.1in
\end{figure}

\noindent\textbf{Rationale Structure.}
We analyze the effect of individual queries designed for rationale extraction. In Figure~\ref{fig:ab2_1}, we evaluate the rationale set generated by the reasoning-policy involving the LLM's self-critic function in~\eqref{eq:2}. The dotted line denotes the performance achieved by employing all $7$ queries, whereas each bar along the x-axis indicates the performance when the $i$-th query is excluded during the dataset construction. Since each query is specifically designed to capture unique features in MDPs, such as goals, state, and return-to-go (illustrated in Figure~\ref{fig:examples}), the exclusion of any one of these queries leads to a performance decline.
In Figure~\ref{fig:ab2_2}, each bar along the x-axis represents the performance achieved when the rationale set is formulated with the queries up to the $i$-th. In the rationale generation process, the $1$st and $2$nd queries encapsulate general information that is applicable to any of the tasks. 
From the $3$rd query onwards, the reasoning becomes increasingly specific to a given task.
Thus, we observe the best performance when all queries are used, building upon the comprehensive information generated in the earlier steps.

\color{black}

\noindent\textbf{Reasoning-policy structure.}
Table~\ref{tab:ab2} shows the effect of our embodied KG and contrastive learning scheme $\mathcal{L}_{\text{Con}}$.
The results indicate that $\ourmodel$, when utilizing both KG and $\mathcal{L}_{\text{Con}}$, achieves the highest performance. 
This is attributed to our embodied KG, which efficiently encapsulates both the evolving embodied information and the agent's interaction experiences. Additionally, in the absence of contrastive learning, the reasoning-policy struggles to extract precise features for the next plan, leading to a performance drop.
We also measure the inference time of $\ourmodel$ and $\ourmodel$ without embodied KG on off-the-shelf devices such as RTX 3090 and 3050 GPUs. As the use of embodied KG allows for more efficient representation, $\ourmodel$ achieves a reduction in inference time by $0.3$ second on average.

\begin{table}[h]
\caption{Ablation on embodied KG and contrastive learning}
\vskip 0.1in
\label{tab:ab2}
\begin{center}
\begin{adjustbox}{width=0.95\linewidth}
\begin{tabular}{l cc c c}
\toprule
\multirow{2}{*}{Method}
& \multicolumn{2}{c}{Inference Time} & \multirow{2}{*}{Seen} & \multirow{2}{*}{Unseen} \\
\cmidrule(rl){2-3}
& RTX $3090$ & RTX $3050$ \\
\midrule
$\ourmodel$ 
&$0.65{\pm0.01}$ & $1.16{\pm0.01}$ & $\mathbf{81.8{\pm0.5}}$ & $\mathbf{46.5{\pm1.0}}$ \\

$\ \ -$ $\mathcal{L}_{\text{Con}}$ 
& - & - & $77.2{\pm0.5}$ & $42.5{\pm1.5}$  \\

$\ \ -$ KG 
& $0.72{\pm0.01}$ & $1.68{\pm0.01}$ & $71.0{\pm6.3}$ & $42.0{\pm6.4}$  \\

$\ \ -$ KG \& $\mathcal{L}_{\text{Con}}$ 
& - & - & $73.2{\pm0.5}$ & $43.3{\pm0.8}$   \\

\bottomrule
\end{tabular}
\end{adjustbox}
\end{center}
\vskip -0.1in
\end{table}

Table~\ref{tab:ab3} shows the effect of our attention structure $\Psi$ consisting $\Psi_\text{c}$ and $\Psi_\text{g}$, which are used in the reasoning-policy $\RPolicy$. In the table, \textit{Iterative} specifies that $\Phi_\text{R}$ is inferred $m$ times sequentially to generate the rationale set without using the attention mechanism. $\Psi_\text{a}$ denotes the basic attention structure~\cite{attn:causal}.
Considering both the success rate and inference time, $\ourmodel$ not only efficiently distills rationales but also offers an effective inference framework for off-the-shelf agents. This is significant when considering that LLM-planner-LLaMA2 reaches $9.37$ seconds at maximum resource usage on an RTX 3090, as in Table~\ref{tab:main}.
\begin{table}[h]
\caption{Ablation on attention structure in reasoning-policy}
\label{tab:ab3}
\vskip 0.1in
\begin{center}
\begin{adjustbox}{width=0.95\linewidth}
\begin{tabular}{l cc cc}
    \toprule
    \multirow{2}{*}{Method}
    & \multicolumn{2}{c}{Inference Time} & \multirow{2}{*}{Seen} & \multirow{2}{*}{Unseen} \\
    \cmidrule(rl){2-3}
    & RTX $3090$ & RTX $3050$ \\
    \midrule
    Iterative
    & $2.92{\pm0.02}$ & $3.08{\pm0.01}$ & $76.0{\pm0.7}$ & $44.6{\pm1.2}$ \\

    $\ourmodel$
    & $0.65{\pm0.01}$ & $1.16{\pm0.01}$ & $\mathbf{81.8{\pm0.5}}$ & $\mathbf{46.5{\pm1.0}}$ \\

    $\ \ -$ $\Psi_\text{c}$
    & $0.63{\pm0.01}$ & $1.16{\pm0.01}$ & $75.9{\pm0.3}$ & $40.6{\pm0.6}$ \\

    $\ \ -$ $\Psi_\text{g}$
    & $0.63{\pm0.01}$ & $1.16{\pm0.01}$ & $76.8{\pm0.3}$ & $40.7{\pm1.4}$ \\

    $\ \ -$ $\Psi_\text{c} \ \& \ \Psi_\text{g}$
    & $0.60{\pm0.01}$ & $1.00{\pm0.01}$ & $63.3{\pm0.1}$ & $34.1{\pm0.1}$ \\

    $\ourmodel$ w $\Psi_\text{a}$
    & $0.63{\pm0.01}$ & $1.17{\pm0.01}$ & $74.7{\pm0.4}$ & $43.6{\pm0.1}$ \\

    \bottomrule
    \end{tabular}
\end{adjustbox}
\end{center}
\vskip -0.1in
\end{table}

\noindent\textbf{sLM Capacity.}
Table~\ref{tab:ab4} shows the performance of $\ourmodel$ with respect to the variations in network parameter sizes for the reasoning-policy $\RPolicy$ and the planning-policy $\DPolicy$.
In our default framework implementation, we utilize the t5-small and gpt2 models for $\Phi_\text{R}$ and $\Phi_\text{D}$, respectively.
The results indicate that the performance improvement is not significant when the parameter size of the planning-policy $\DPolicy$ increases.
In contrast, enhancing the parameter size of the reasoning-policy $\RPolicy$ results in performance gains, showing an average increase of $6.57\%$ when comparing the t5-small and the t5-large used for $\RPolicy$ in unseen settings.
Specifically, the smaller sLM (t5-small) tends to overfit on the training datasets, which might yield better performance in the Seen category compared to the mid-sized sLM (t5-base). For the larger sLM (t5-large), a performance improvement is noted in the Seen category, attributed to its enhanced reasoning capabilities. In contrast, the Unseen settings demonstrate a linear performance increase as the parameter size of the reasoning policy grows, suggesting that a larger parameter size significantly boosts the generalization ability of the model.
This indicates the benefits of distilling rationales from LLMs, which plays a crucial role in establishing a robust sLM-based policy. 

\begin{table}[h]
\caption{$\ourmodel$ performance w.r.t. policy network sizes}
\vskip 0.1in
\label{tab:ab4}
\begin{center}
\begin{adjustbox}{width=0.95\linewidth}
\begin{tabular}{llccc}
\toprule
$\Phi_\text{R}$ & $\Phi_\text{P}$ & Parameter Size & Seen & Unseen \\
\midrule
t5-small & gpt2         & $0.06$B+$0.1$B    & $81.8{\pm0.5}$ & $46.5{\pm1.0}$ \\
t5-small & gpt2-medium  & $0.06$B+$0.4$B    & $81.4{\pm0.1}$ & $46.1{\pm0.5}$ \\
t5-small & gpt2-large   & $0.06$B+$0.8$B    & $81.8{\pm0.1}$ & $46.1{\pm1.1}$ \\

t5-base & gpt2          & $0.2$B+$0.1$B   & $79.4{\pm0.5}$ & $48.5{\pm1.2}$ \\
t5-base & gpt2-medium   & $0.2$B+$0.4$B   & $78.8{\pm0.6}$ & $48.6{\pm0.6}$ \\
t5-base & gpt2-large    & $0.2$B+$0.8$B   & $80.0{\pm1.3}$ & $49.0{\pm1.5}$ \\

t5-large & gpt2         & $0.7$B+$0.1$B   & $82.1{\pm1.0}$ & $52.5{\pm0.7}$ \\
t5-large & gpt2-medium  & $0.7$B+$0.4$B   & $81.8{\pm0.6}$ & $52.9{\pm0.9}$ \\
t5-large & gpt2-large   & $0.7$B+$0.8$B   & $81.2{\pm0.4}$ & $53.3{\pm0.7}$ \\
\bottomrule
\end{tabular}
\end{adjustbox}
\end{center}
\vskip -0.1in
\end{table}

\section{Conclusion}
\label{sec:con}

We introduced $\ourmodel$, a novel framework that effectively distills the reasoning capabilities of LLMs into more compact sLMs for executing complex embodied tasks in device-constrained environments. The framework operates in a strategic distillation process, involving embodied rational data construction from an LLM, data-driven embodied policy distillation to an sLM, and task planning with the sLM.
This allows for the efficient use of LLM-powered complex task planning functions in real-world time-constrained settings while ensuring the adaptability to resource-constrained agent conditions through two-step distillation into reasoning and decision-making. 

\noindent\textbf{Limitation.}
As $\ourmodel$ employs pre-trained sLMs, there is a potential dependency on the pre-trained knowledge embedded in the sLMs. In Table~\ref{tab:ab4}, we observe that a reduced network capacity of sLMs leads to decreased performance in unseen settings. This indicates that the limited network capacity of the sLM hinders the distillation of reasoning capabilities, consequently affecting the zero-shot adaptation in environments with significant domain shifts.

\noindent\textbf{Future Work.}
Future directions for our research include enhancing the framework's ability for few-shot optimization, especially in scenarios with significant domain shifts, aiming to explore the versatility of LLMs.

\section*{Acknowledgements}
We would like to thank anonymous reviewers for their valuable comments and suggestions.
This work was supported by the Institute of Information \& Communications Technology Planning \& Evaluation (IITP) grant funded by the Korea government (MSIT) (No. 2022-0-01045, No. 2022-0-00043, No. 2019-0-00421, No. 2020-0-01821), by the National Research Foundation of Korea (NRF) grant funded by MSIT (No. RS-2023-00213118), and by Samsung Electronics.

\section*{Impact Statement}
This paper presents work whose goal is to advance the field of 
Machine Learning. There are many potential societal consequences 
of our work, none which we feel must be specifically highlighted here.


\bibliography{main_icml}
\bibliographystyle{icml2024}

%%%%%%%%%%%%%%%%%%%%%%%%%%%%%%%%%%%%%%%%%%%%%%%%%%%%%%%%%%%%%%%%%%%%%%%%%%%%%%%
%%%%%%%%%%%%%%%%%%%%%%%%%%%%%%%%%%%%%%%%%%%%%%%%%%%%%%%%%%%%%%%%%%%%%%%%%%%%%%%
% APPENDIX
%%%%%%%%%%%%%%%%%%%%%%%%%%%%%%%%%%%%%%%%%%%%%%%%%%%%%%%%%%%%%%%%%%%%%%%%%%%%%%%
%%%%%%%%%%%%%%%%%%%%%%%%%%%%%%%%%%%%%%%%%%%%%%%%%%%%%%%%%%%%%%%%%%%%%%%%%%%%%%%

\appendix
\renewcommand{\theequation}{A.\arabic{equation}}
\renewcommand{\thefigure}{A.\arabic{figure}}
\renewcommand{\thetable}{A.\arabic{table}}
\onecolumn

\section{Environment settings}

\subsection{ALFRED}
\begin{figure}[h]
    \centering
    \subfigure[Example of Heat \& Place task.]{
        \centering
        \includegraphics[width=0.46\linewidth]{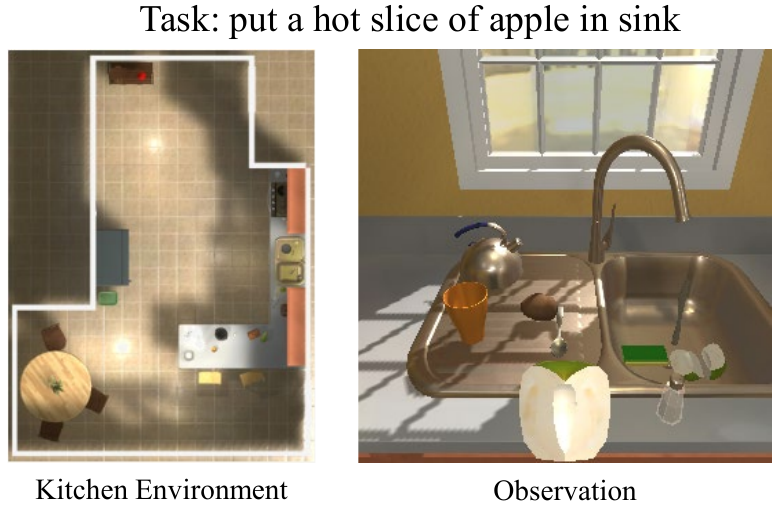}
    }
    % \hspace{10pt}
    \subfigure[Example of Pick Two \& Place task.]{
        \centering
        \includegraphics[width=0.46 \linewidth]{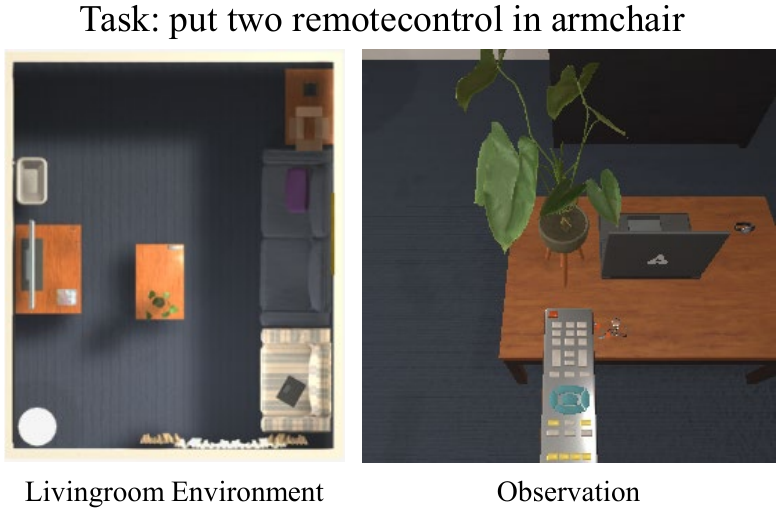}
    }
    
    \centering
    \subfigure[Example of Examine \& in Light task.]{
        \centering
        \includegraphics[width=0.46\linewidth]{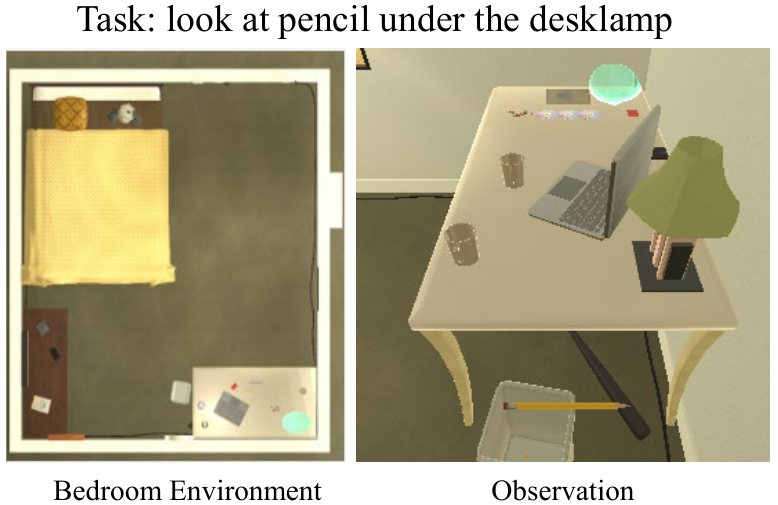}
    }
    % \hspace{10pt}
    \subfigure[Example of Pick \& Place task.]{
        \centering
        \includegraphics[width=0.46 \linewidth]{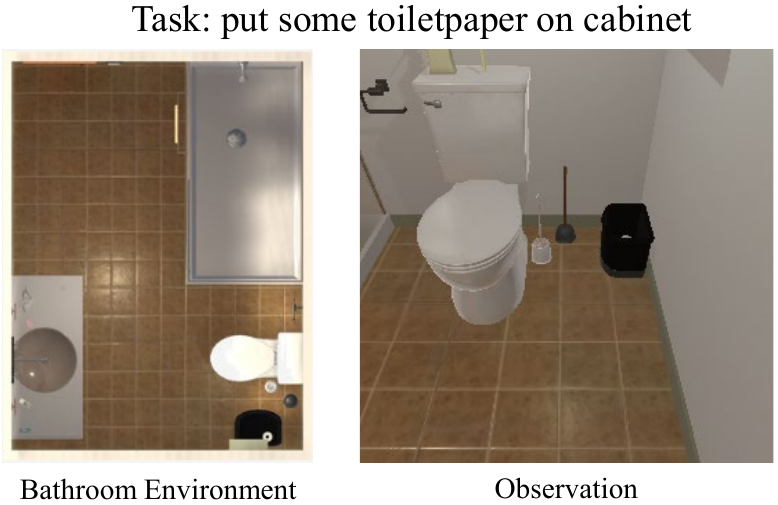}
    }
    \label{app:fig:1}
    \caption{Task examples set within different indoor scenes. The observation includes a variety of objects with which the agent can interact and alter states to complete the given task.}
\end{figure}

We utilize ALFRED~\cite{ben:ALFRED20}, which provides comprehensive vision-and-language navigation and rearrangement tasks for embodied AI. 
This environment requires an agent to follow language formatted instructions to accomplish real-world-like household tasks.
ALFRED features 58 different object types (e.g., bread) and 26 receptacle types (e.g., plate) across 120 various indoor scenes (e.g., kitchen).
It supports 4703 unique tasks, each configured by combining these elements with one of 7 instruction types (e.g., pick \& place), such as “Put a keychain in a plate and then put them on a shelf”.
This complexity and diversity makes ALFREDD an ideal benchmark for evaluating models that emphasize hierarchy, modularity, and advanced reasoning and planning capabilities.
The detail of instructions and excutable plans are listed in Table~\ref{app:tab1}.
Furthermore, the visualizations of various indoor scenes and observations in ALFRED are shown in Figure~\ref{app:fig:1}

\begin{table}[h]
\caption{Instructions and executable plans in ALFRED environment}
\vskip 0.1in
\label{app:tab1}
\begin{center}
\begin{small}
\begin{tabular}{l ll}

\toprule
& \textbf{Type} & \textbf{Example} \\
\midrule
\multirow{7}{*}{Instructions}  & Pick \& Place & Put a watch on a table. \\
                        & Stack \& Place & Put a bowl with a spoon in it on the table. \\
                        & Pick Two \& Place & Put two pencils in a drawer. \\
                        & Clean \& Place & Put a clean rag on the top shelf of a barred rack. \\
                        & Heat \& Place & Put a cooked potato slice on the counter \\
                        & Cool \& Place & Put a slice of cold lettuce on a counter. \\
                        & Examine \& in Light & Pick up a book and turn on a lamp. \\
\midrule
\multirow{11}{*}{Plans} & OpenObject {[}Object{]} & OpenObject GarbageCan \\
                        & CloseObject {[}Object{]} & CloseObject GarbageCan \\
                        & ToggleObject {[}Object{]} & ToggleObject FloorLamp \\ 
                        & SliceObject {[}Object{]} & SliceObject Potato \\
                        & GotoLocation {[}Receptacle Object{]} & GotoLocation SideTable \\
                        & PickupObject {[}Object{]} {[}Receptacle Object{]} & PickupObject ButterKnife SideTable \\ 
                        & PutObject {[}Object{]} {[}Receptacle Object{]} & PutObject Pan DiningTable \\ 
                        & CoolObject {[}Object{]} {[}Receptacle Object{]} & CoolObject Apple Fridge \\ 
                        & HeatObject {[}Object{]} {[}Receptacle Object{]} & HeatObject Mug Microwave \\ 
                        & CleanObject {[}Object{]} {[}Receptacle Object{]} & CleanObject Tomato Sink \\
                        & End & End \\ 
\bottomrule

\end{tabular}
\end{small}
\end{center}
\vskip -0.1in
\end{table}

\subsection{Expert Dataset}

\textbf{Expert Dataset and Evaluation Task Settings.}
To generate an expert dataset, we use planning domain definition language rules~\cite{app:pddl}. For implementation, we use the open source project\footnote{https://github.com/askforalfred/alfred}.
We collect $312$ expert trajectories in a variety of tasks varying the starting positions of the agent and objects as well as the indoor scenes.

We organize the evaluation tasks into $4$ distinct categories based on task similarities with the expert dataset: \textbf{Train}, \textbf{Seen}, \textbf{Unseen Spatial}, and \textbf{Unseen Environment}.
For the \textbf{Train} category, the tasks are identical to those tasks in the expert dataset.
The \textbf{Seen} category maintains the same tasks as in the expert dataset, but task-irrelevant objects are randomly positioned at the start. For this, we evaluate $528$ tasks.
In the \textbf{Unseen Spatial} category, all objects are placed randomly, and tasks are either defined by new task descriptions or optimal planning sequences not included in the Train category. For this, we evaluate $1415$ tasks.
Lastly, for the most challenging category \textbf{Unseen Environment} where all objects are randomly placed, and the task or indoor scenes are not presented in the Train category. We utilize $58$ tasks for evaluating the Unseen Environment category.
For all models that require training, we conduct evaluations using three distinct seeds and report their average performance, along with the associated variances.

\section{Implementation Details}
In this section, we provide the implementation details of our proposed framework $\ourmodel$ and each comparison. 
Our framework is implemented using Python v3.9 and PyTorch v2.0.1, trained on a system of an Intel(R) Core (TM) i9-10980XE processor and an NVIDIA RTX A6000 GPU.
For comparisons, we implement $3$ types of widely used approaches: language planning, knowledge distillation, and end-to-end methodologies.

\subsection{Language Planning Approach}
For the language planning approaches, we employ $3$ different methodologies: SayCan, LLM-planner, and ZSP~\cite{llmagent:saycan, llmagent:llmplanner, llmagent:zsp}. For generating high-level plans, we utilize various LMs like PaLM, LLAMA, and GPT2-large.

\textbf{SayCan}~\cite{llmagent:saycan} integrates pretrained skills with language models, generating plans that are feasible to the context.
SayCan achieves this by combining affordance scores derived from the LM with the agent's experiences.
In line with SayCan's methodology, we calculate embodied affordance scores by utilizing object presence information.
For implementation, we refer to the open source project\footnote{https://github.com/google-research/google-research/tree/master/saycan}.

\textbf{ZSP}~\cite{llmagent:zsp} leverages the capabilities of the LLMs for embodied task planning by interpreting high-level task descriptions and formulating sequential strategies, thus efficiently performing embodied tasks. ZSP accomplishes this by crafting step-by-step prompts based on examples of similar successful tasks, followed by sampling executable plans using the LLM in conjunction with these provided examples.
For implementation, we refer to the open source project\footnote{https://github.com/huangwl18/language-planner}. 

\textbf{LLM-planner}~\cite{llmagent:llmplanner} leverages the LLMs for few-shot planning, empowering embodied agents to perform complex tasks in environments with observed information, guided by natural language instructions.
For implementation, we refer to the open source\footnote{https://github.com/OSU-NLP-Group/LLM-Planner/}. 

In text generation configuration, temperature controls the degree of randomness in the generation process. A lower temperature results in more predictable and consistent text, while a higher temperature can produce more diverse and sometimes unexpected outcomes. Top $k$ sampling limits the model to consider only the top $k$ most probable next words when choosing the next word in the sequence. This method helps to constrain randomness, thereby enhancing the quality of the generated text. Top $p$ involves selecting the smallest set of words whose cumulative probability exceeds $p$ for choosing the next word.

The hyperparameter settings for language planning approaches are summarized in Table~\ref{hyper:lp}. 

\begin{table}[h]
      \caption{Hyperparamer settings for language planning approaches}
      \vskip 0.1in
      \centering
      % \adjustbox{width=\linewidth}{
        \label{hyper:lp}
        \begin{tabular}{lcccr}
        \toprule
        \textbf{Hyperparameters} & \textbf{Value} \\
        \midrule
        \multicolumn{2}{l}{\underline{\textbf{LLM Configuration}}} \\
        PaLM    & text-bison-001 \\
        LLaMA2  & llama-2-7b \\
        GPT2    & gpt2-large \\
        \multirow{2}{*}{In-Context Example Retriever}
        & paraphrase-MiniLM-L6-v2 (LLM-Planner) \\
        & stsb-roberta-large (ZSP) \\
        Number of Prompts & 4 (LLM-Planner, ZSP) \\ 
        \midrule
        \multicolumn{2}{l}{\underline{\textbf{Text Generation Configuration}}} \\
        % \midrule
        Sampling Method & beam search \\
        Beam Size & $3$ \\
        Temperature & $0.01$ \\
        Top $k$ & $5$ \\
        Top $p$ & $0.3$ \\
        Maximum New Tokens & $40$ \\
        \bottomrule
        \end{tabular}
    % }
    \vskip -0.1in
\end{table}

\subsection{Knowledge Distillation Approach}
For the knowledge distillation approaches, we employ two different algorithms: SCoTD and SCOTT~\cite{cot:symbolic, cot:scott}. For distilling the reasoning-policy to produce MDP-featured rationales, we implement each method to create the rationale dataset accordingly.

\textbf{SCoTD}~\cite{cot:symbolic} is a CoT distillation method to train an sLM. It utilizes a LLM to generate a variety of rationales with answers, which are then used to educate the sLM.
We employ SCoTD to generate and train rationale data for the reasoning-policy, and then utilize the distilled rationales to further train the planning-policy.
% The hyperparameter settings are summarized in Table~\ref{hyper:SCoTD}.

\textbf{SCOTT}~\cite{cot:scott} is a consistency knowledge distillation method to train a smaller, self-consistent CoT model from a much larger teacher model. SCOTT uses contrastive decoding to elicit better rationale supervision and a counterfactual reasoning objective to align the student model's predictions with these rationales.
We utilize SCOTT for creating rationale data and subsequently training the reasoning-policy. The learned rationales from reasoning-policy are then applied to train the planning-policy.
For implementation, we refer to the open source project\footnote{https://github.com/wangpf3/consistent-CoT-distillation}.

The hyperparameter settings for knowledge distillation approaches are summarized in Table~\ref{hyper:kd}.

\begin{table}[H]   
    \centering
      % \adjustbox{width=\linewidth}{
        \centering
        \caption{Hyperparamer settings for knowledge distllation approaches}
        \vskip 0.1in
        \label{hyper:kd}
        \begin{tabular}{lcccr}
        \toprule
        \textbf{Hyperparameters} & \textbf{Value} \\
        \midrule
        \multicolumn{2}{l}{\underline{\textbf{Source LLM}}} \\
        PaLM            & text-bison-001 \\
        Temperature     & $0.7$ (SCoTD), $0.1$ (SCOTT)\\
        Return samples  & $3$ (SCoTD), $1$ (SCOTT) \\
        \midrule
        \multicolumn{2}{l}{\underline{\textbf{Reasoning-policy}}} \\
        sLM             & t5-small \\
        Train epochs    & $100$ \\
        Batch size      & $1$ \\ 
        Optimizer       & SGD \\
        Learning rate   & $5$e$-5$ \\
        \midrule
        \multicolumn{2}{l}{\underline{\textbf{Planning-policy}}} \\
        sLM             & gpt2 \\
        Train epochs    & $20$ \\
        Batch size      & $2$ \\ 
        Optimizer       & SGD \\
        Learning rate & $3$e$-5$ \\
        \midrule
        \multicolumn{2}{l}{\underline{\textbf{Text Generation Configuration}}} \\
        % \midrule
        Sampling Method & beam search \\
        Beam Size & $3$ \\
        Temperature & $0.01$ \\
        Top $k$ & $5$ \\
        Top $p$ & $0.3$ \\
        Maximum New Tokens & $40$ \\
        \bottomrule
        \end{tabular}
    % }
    \vskip -0.1in
\end{table}

\subsection{End2End}
End2End~\cite{lmagent:butlers} is a method for embodied task planning that specifically utilizes the GPT-2 model, trained with direct supervision on expert data. This approach forms the foundational backbone for our planning policy $\Phi_\text{P}$ implementation.
For implementation, we refer to the open source project\footnote{https://github.com/vmicheli/lm-butlers}. 
The hyperparameter settings for End2End are summarized in Table~\ref{hyper:e2e}.

\begin{table}[h]   
        \centering
        \caption{Hyperparamer settings for End2End}
        \vskip 0.1in
        \label{hyper:e2e}
        \begin{tabular}{lcccr}
        \toprule
        \textbf{Hyperparameters} & \textbf{Value} \\
        \midrule    
        sLM             & gpt2 \\
        Train epochs    & 100 \\
        Batch size      & 1 \\ 
        Optimizer       & SGD\\
        Learning rate & $3$e$-5$ \\
        \bottomrule
    \end{tabular}
    \vskip -0.1in
\end{table}

\subsection{$\ourmodel$}
The entire procedure of our $\ourmodel$ consists of rationale dataset construction and policy distillation via embodied knowledge graph phases.

\subsubsection{Rationale Dataset Construction}
In the rationale dataset construction phase, we use PaLM~\cite{palm} as the source LLM, exploiting its reasoning capabilities. We formulate $7$ queries to extract rationales from the LLM and manually design $9$ initial examples of query-rationale pairs for each expert transition. To calculate similarity between language embeddings of $\tau$ and $c$, we use contextual embedding model,
To utilize the LLM as a critic function, we query the LLM to assess whether the generated rationales are sufficient to generate the plan. For instance, we ask, `Can the rationale \{\textit{rationale}\} lead to the next plan \{\textit{plan}\}? Answer with yes or no.'. To ensure accurate evaluations, we pose several variations on the critic prompt and determine the final critic score based on majority voting.

Algorithm~\ref{alg:gen} lists the dataset construction procedures.

\begin{algorithm}[t]
\caption{Rationale Dataset Construction}
\label{alg:gen}
Expert Dataset $\Dataset = \{\tau_i\}_{i}$, LLM $\LLM$

\begin{algorithmic}[1]
\STATE{Initialize rationale dataset, $\CoTDataset \leftarrow \emptyset$}
\FOR{$\tau_i=(o,a,h)$ in $\Dataset$}
    \WHILE{true}
        \STATE{Sample a set of tuples, $\mathcal{C}=\{c_1,...,c_n\} \sim \CoTDataset$}
        \STATE{Retrieve in-context example set, $\mathcal{C}_k = F(\tau, \mathcal{C})$}
        \STATE{Initialize rationale set, $\mathcal{R} \leftarrow \emptyset$}
        \FOR{$j = 1, ..., m$}
            \STATE{Generate rationale $r_j$ given query $q_j$ and~\eqref{eq:1}}
            \STATE{Update $\mathcal{R} \leftarrow \mathcal{R} \cup \{r_j\}$}
        \ENDFOR
        \STATE{Construct new tuple $c = (o, a, h, \mathcal{R})$}
        \IF{$c$ passes the self-critic using~\eqref{eq:2}}
            \STATE{$\CoTDataset \leftarrow \CoTDataset \cup \{c\}$}
            \STATE{\textbf{break}}
        \ENDIF
    \ENDWHILE
\ENDFOR
\STATE{\textbf{return} $\CoTDataset=\{c_i\}_{i}$}
\end{algorithmic}
\end{algorithm}

\subsubsection{Policy Distillation via Embodied Knowledge Graph}
In the policy distillation phase, we distill the reasoning capabilities from the LLM into an sLM-based policy $\sLM$, which is structured with a two-tier hierarchy consisting of the reasoning-policy $\RPolicy$ and the planning-policy $\DPolicy$.

\textbf{Reasoning-policy.}
For the reasoning-policy $\RPolicy$, we utilize a pre-trained language model with an encoder-decoder structure, specifically t5-small~\cite{t5}, as our default setting. 
The dimension of prefix prompts $\theta_\text{Pre}^{(i)}$, postfix prompts $\theta_\text{Pos}^{(i)}$ and decoder prompts $\theta_\text{Dec}^{(i)}$ are set to be $20$. 
Our implementation of the attention module $\Psi$ incorporates two distinct attention mechanisms: causal attention and gated attention, each comprising a single attention layer.
The causal attention module uses a causal mask, while the gated attention module includes an additional learnable gate function.
$\RPolicy$ is optimize by \eqref{loss:RationaleGenerationLoss} and \eqref{loss:ContrastiveLoss}.

\textbf{Planning-policy.}
For the planning-policy $\DPolicy$, we utilize a pre-trained language model with a decoder structure, specifically gpt2~\cite{gpt2}, as our default setting. 
$\DPolicy$ is optimize by \eqref{loss:PlanGenerationLoss}. 

The hyperparameter settings are summarized in Table~\ref{hyper:deder}. 

\begin{table}[h]   
    \centering
      % \adjustbox{width=\linewidth}{
        \centering
        \caption{Hyperparamer settings for $\ourmodel$}
        \vskip 0.1in
        \label{hyper:deder}
        \begin{tabular}{lc|lc}
        \toprule
        \textbf{Hyperparameters} & \textbf{Value} & \textbf{Hyperparameters} & \textbf{Value} \\
        \midrule
        \multicolumn{2}{l|}{\underline{\textbf{Source LLM}}} & \multicolumn{2}{l}{\underline{\textbf{Planning-policy}}} \\
        PaLM            & text-bison-001 & sLM             & gpt2 \\
        Temperature     & $0.1$ & Train epochs    & $20$\\
        Return samples  & $1$ & Batch size      & $2$ \\
        \cmidrule{1-2}
        \multicolumn{2}{l|}{\underline{\textbf{Reasoning-policy}}} & Optimizer       & SGD \\
        sLM             & t5-small & Learning rate & $3$e$-5$\\
        \cmidrule{3-4}
        Encoder prompt length & 20 ($\theta_\text{Pre}^{(i)}$), 20 ($\theta_\text{Pos}^{(i)}$) & \multicolumn{2}{l}{\underline{\textbf{Text Generation Configuration}}} \\
        Decoder prompt length & 20 ($\theta_\text{Dec}^{(i)}$) & Sampling Method & beam search \\
        Train epochs    & $100$ & Beam Size & $3$ \\
        Batch size      & $1$ & Temperature & $0.01$ \\
        Optimizer       & SGD  &  Top $k$ & $5$  \\
        Learning rate   & $5$e$-5$ &  Top $p$ & $0.3$  \\
        Scaling factor $\alpha$ & 0.5 & Maximum New Tokens & $40$ \\
        \bottomrule
        \end{tabular}
    % }
    \vskip -0.1in
\end{table}

\section{Additional Experiments}
For further investigation, we report additional experimental results.

\subsection{Details of Ablation on Rationale Dataset Construction}
Table~\ref{app:tab:ab1} shows detailed experiment results of Table~\ref{tab:ab1}.
$\ourmodel$ consistently demonstrates improved performance than the few-shot CoT method across various language models.
The lower performance is observed when using GPT2 for rationale extraction, demonstrating the limited reasoning capability of sLM with in-context learning. 
The slight performance difference between the Few-shot approach and $\ourmodel$, when using GPT3.5's chat-based architecture, can be attributed to its conversational design focus.
For the optimal application of MDP-featured in-context learning with a Chat LLM, distinct from a text generation model, crafting an in-context example design specifically for dialogue interactions becomes crucial.

\begin{table*}[h]
\caption{Details of ablation on rationale dataset construction}
\vskip 0.1in
\begin{center}
\begin{adjustbox}{width=0.95\textwidth}
    \begin{tabular}{l c cc cc cc cc}
    \toprule
    \multirow{2}{*}{Method} & \multirow{2}{*}{LM}
    & \multicolumn{2}{c}{Train} & \multicolumn{2}{c}{Seen} & \multicolumn{2}{c}{Unseen Spatial} & \multicolumn{2}{c}{Unseen Environment} \\
    
    \cmidrule(rl){3-4} \cmidrule(rl){5-6} \cmidrule(rl){7-8} \cmidrule(rl){9-10}
     & & SR  & GC  & SR  & GC  & SR  & GC  & SR  & GC  \\
    \midrule
    
    Few-shot
    & GPT2
    & $41.9{\pm19.3}$ & $67.4{\pm13.5}$
    & $2.4{\pm0.5}$ & $23.1{\pm0.7}$
    & ${0.6\pm0.1}$ & $22.1{\pm0.9}$
    & $0.0{\pm0.0}$ & ${20.5\pm1.2}$ \\

    $\ourmodel$
    & GPT2
    & $60.8{\pm4.0}$ & $82.4{\pm2.2}$
    & $53.5{\pm2.2}$ & $75.3{\pm1.3}$
    & $27.3{\pm61.7}$ & $61.7{\pm0.7}$
    & $19.4{\pm15.5}$ & $49.1{\pm2.8}$ \\

    \midrule

    Few-shot
    & GPT3
    & $100.0{\pm0.0}$ & $100.0{\pm0.0}$
    & $72.8{\pm0.1}$ & $87.4{\pm0.2}$
    & $48.6{\pm0.4}$ & $79.0{\pm0.1}$
    & $28.7{\pm2.6}$ & $62.9{\pm2.1}$ \\

    $\ourmodel$
    & GPT3
    & $100.0{\pm0.0}$ & $100.0{\pm0.0}$
    & $72.8{\pm0.1}$ & $88.8{\pm0.2}$
    & $46.9{\pm0.5}$ & $77.3{\pm0.4}$
    & $37.4{\pm2.0}$ & $65.3{\pm0.9}$ \\

    Few-shot
    & GPT3.5
    & $100.0{\pm0.0}$ & ${100.0\pm0.0}$
    & $78.3{\pm0.6}$ & $90.5{\pm0.2}$
    & $52.0{\pm0.7}$ & $80.0{\pm0.3}$
    & $39.7{\pm0.0}$ & $66.5{\pm0.3}$ \\

    $\ourmodel$
    & GPT3.5
    & $100.0{\pm0.0}$ & $100.0{\pm0.0}$
    & $78.4{\pm1.4}$ & $91.5{\pm2.3}$
    & $50.0{\pm0.6}$ & $80.3{\pm0.2}$
    & $39.9{\pm2.3}$ & $68.6{\pm1.4}$ \\

    Few-shot
    & PaLM
    & $100.0{\pm0.0}$ & ${100.0\pm0.0}$
    & $76.9{\pm0.3}$ & $89.7{\pm0.1}$
    & $49.3{\pm0.3}$ & $79.7{\pm0.2}$
    & $29.9{\pm1.0}$ & $63.5{\pm0.4}$ \\

    $\ourmodel$
    & PaLM
    & $100.0{\pm0.0}$ & $100.0{\pm0.0}$
    & $\mathbf{81.8{\pm0.5}}$ & $\mathbf{92.2{\pm0.2}}$
    & $\mathbf{52.7{\pm1.0}}$ & $\mathbf{81.2{\pm0.4}}$
    & $\mathbf{40.3{\pm0.9}}$ & $\mathbf{68.7{\pm0.6}}$ \\
    
    \bottomrule
    \end{tabular}

\end{adjustbox}
\end{center}
\label{app:tab:ab1}
\vskip -0.1in
\end{table*}

\subsection{Details of Embodied Knowledge Graph and Contrastive Learning}
Table~\ref{app:tab:ab2} shows detailed experiment results of Table~\ref{tab:ab2}. 
We measured inference times several off-the-shelf devices such as RTX 3090, 3050 and 2080 Ti GPUs. 
$\ourmodel$ ensures real-time inference speeds across these various devices while consistently yielding superior performance compared to other ablated comparisons.

\begin{table*}[h]
\caption{Details of ablation on embodied KG and contrastive learning}
\vskip 0.1in
\begin{center}
\begin{adjustbox}{width=0.95\textwidth}
    \begin{tabular}{l ccc cc cc cc cc}
    \toprule
    \multirow{2}{*}{Method} & \multicolumn{3}{c}{Inference Time}
    & \multicolumn{2}{c}{Train} & \multicolumn{2}{c}{Seen} & \multicolumn{2}{c}{Unseen Spatial} & \multicolumn{2}{c}{Unseen Environment} \\
    
    \cmidrule(rl){2-4} \cmidrule(rl){5-6} \cmidrule(rl){7-8} \cmidrule(rl){9-10} \cmidrule(rl){11-12}
     & RTX 3090 & RTX 3050 & RTX 2080 Ti & SR  & GC  & SR  & GC  & SR  & GC  & SR  & GC  \\
    \midrule
    
    $\ourmodel$
    % & $0.65{\pm0.01}$($1.54{\pm0.01}$) & ${0.77{\pm0.02}}$($1.3{\pm0.03}$)
    & $0.65{\pm0.01}$ & $1.16{\pm0.01}$ & ${0.77{\pm0.02}}$
    & $100.0{\pm0.0}$ & $100.0{\pm0.0}$
    & $\mathbf{81.8{\pm0.5}}$ & $\mathbf{92.2{\pm0.2}}$
    & $\mathbf{52.7{\pm1.0}}$ & $\mathbf{81.2{\pm0.4}}$
    & $\mathbf{40.3{\pm0.9}}$ & $\mathbf{68.7{\pm0.6}}$ \\

    $\ \ -$ $\mathcal{L}_{\text{Con}}$
    % & $0.65{\pm0.01}$($1.54{\pm0.01}$) & ${0.77{\pm0.02}}$($1.3{\pm0.03}$)
    & $0.65{\pm0.01}$ & $1.16{\pm0.01}$ & ${0.77{\pm0.02}}$
    & $100.0{\pm0.0}$ & $100.0{\pm0.0}$
    & $77.2{\pm0.5}$ & $89.9{\pm0.3}$
    & $51.1{\pm1.0}$ & $80.3{\pm0.4}$
    & $33.9{\pm2.0}$ & $65.2{\pm1.3}$ \\

    $\ \ -$KG
    % & $0.72{\pm0.01}$($1.39{\pm0.01}$) & ${1.12\pm0.01}$($0.90{\pm0.09}$)
    & $0.72{\pm0.01}$ & $1.68{\pm0.01}$ & ${1.12\pm0.01}$
    & $99.9{\pm0.1}$ & $99.9{\pm0.1}$
    & $71.0{\pm6.3}$ & $87.1{\pm3.5}$
    & $48.3{\pm7.2}$ & $78.4{\pm4.0}$
    & $35.6{\pm5.6}$ & $65.7{\pm3.9}$ \\

    $\ \ -$KG \& $\mathcal{L}_{\text{Con}}$
    % & $0.72{\pm0.01}$($1.39{\pm0.01}$) & ${1.12\pm0.01}$($0.90{\pm0.09}$)
    & $0.72{\pm0.01}$ & $1.68{\pm0.01}$ & ${1.12\pm0.01}$
    & $93.7{\pm1.6}$ & $97.5{\pm0.8}$
    & $73.2{\pm0.5}$ & $87.8{\pm0.4}$
    & $49.5{\pm0.2}$ & $79.9{\pm0.2}$
    & $37.1{\pm1.2}$ & $66.7{\pm0.0}$ \\
    
    \bottomrule
    \end{tabular}

\end{adjustbox}
\end{center}
\label{app:tab:ab2}
\vskip -0.1in
\end{table*}

\subsection{Details of Reasoning-policy Structure}
Table~\ref{app:tab:ab3} shows detailed experiment results of Table~\ref{tab:ab3}.
$\ourmodel$ ensures real-time inference speeds across various off-the-shelf devices while maintaining consistently superior performance compared to other ablation comparisons.

\begin{table*}[h]
\caption{Details of reasoning-policy structure}
\vskip 0.1in
\begin{center}
\begin{adjustbox}{width=0.95\textwidth}
    \begin{tabular}{l ccc cc cc cc cc}
    \toprule
    \multirow{2}{*}{Method} & \multicolumn{3}{c}{Inference Time}
    & \multicolumn{2}{c}{Train} & \multicolumn{2}{c}{Seen} & \multicolumn{2}{c}{Unseen Spatial} & \multicolumn{2}{c}{Unseen Environment} \\
    
    \cmidrule(rl){2-4} \cmidrule(rl){5-6} \cmidrule(rl){7-8} \cmidrule(rl){9-10} \cmidrule(rl){11-12}
     & RTX 3090 & RTX 3050 & RTX 2080 Ti & SR  & GC  & SR  & GC  & SR  & GC  & SR  & GC  \\
    \midrule

    Iterative
    % & $2.22{\pm0.02}$($0.46{\pm0.01}$) & $2.5{\pm0.01}$($0.40{\pm0.01}$)
    & $2.92{\pm0.02}$ & $3.08{\pm0.01}$ & $3.2{\pm0.01}$
    & $100.0{\pm0.0}$ & $100.0{\pm0.0}$
    & $76.0{\pm0.7}$ & $89.8{\pm0.2}$
    & $\mathbf{54.0{\pm0.4}}$ & $\mathbf{81.4{\pm0.2}}$
    & $35.1{\pm2.0}$ & $65.9{\pm1.7}$ \\
    
    $\ourmodel$
    % & $0.65{\pm0.01}$($1.54{\pm0.01}$) & ${0.77{\pm0.02}}$($1.3{\pm0.03}$)
    & $0.65{\pm0.01}$ & $1.16{\pm0.01}$ & ${0.77{\pm0.02}}$
    & $100.0{\pm0.0}$ & $100.0{\pm0.0}$
    & $\mathbf{81.8{\pm0.5}}$ & $\mathbf{92.2{\pm0.2}}$
    & $52.7{\pm1.0}$ & $81.2{\pm0.4}$
    & $\mathbf{40.3{\pm0.9}}$ & $\mathbf{68.7{\pm0.6}}$ \\

    $\ \ - \Psi_\text{c}$
    % & $0.63{\pm0.01}$($1.58{\pm0.01}$) & $0.74{\pm0.01}$($1.35{\pm0.01}$)
    & $0.63{\pm0.01}$ & $1.47{\pm0.01}$ & $0.74{\pm0.01}$
    & $99.9{\pm0.1}$ & $99.9{\pm0.0}$
    & $75.9{\pm0.3}$ & $89.6{\pm0.1}$
    & $52.2{\pm0.2}$ & $80.7{\pm0.1}$
    & $29.9{\pm1.0}$ & $63.8{\pm0.5}$ \\

    $\ \ - \Psi_\text{g}$
    % & $0.63{\pm0.01}$($1.58{\pm0.01}$) & $0.74{\pm0.01}$($1.35{\pm0.01}$)
    & $0.63{\pm0.01}$ & $1.16{\pm0.01}$ & $0.74{\pm0.01}$
    & $100.0{\pm0.0}$ & $100.0{\pm0.0}$
    & $76.8{\pm0.3}$ & $89.6{\pm0.1}$
    & $50.9{\pm0.1}$ & $79.3{\pm0.1}$
    & $30.5{\pm2.6}$ & $62.9{\pm1.1}$ \\

    $\ \ - \Psi_\text{c} \& \Phi_\text{g}$
    % & $0.60{\pm0.01}$($1.66{\pm0.01}$) & $0.64{\pm0.09}$($1.57{\pm0.20}$)
    & $0.60{\pm0.01}$ & $1.00{\pm0.01}$ & $0.64{\pm0.09}$
    & $100.0{\pm0.0}$ & $100.0{\pm0.0}$
    & $63.3{\pm0.1}$ & $83.1{\pm0.1}$
    & $39.5{\pm0.4}$ & $73.5{\pm0.4}$
    & $28.7{\pm1.0}$ & $61.2{\pm0.4}$ \\

    $\ourmodel$ w $\Psi_\text{a}$
    % & $0.63{\pm0.01}$($1.58{\pm0.1}$) & $0.75{\pm0.01}$($1.34{\pm0.02}$)
    & $0.63{\pm0.01}$ & $1.16{\pm0.01}$ & $0.75{\pm0.01}$
    & $99.9{\pm0.1}$ & $99.9{\pm0.0}$
    & $74.7{\pm0.4}$ & $88.6{\pm0.2}$
    & $52.2{\pm0.2}$ & $80.7{\pm0.1}$
    & $35.1{\pm1.0}$ & $64.1{\pm0.8}$ \\

    \bottomrule
    \end{tabular}

\end{adjustbox}
\end{center}
\label{app:tab:ab3}
\vskip -0.1in
\end{table*}

\subsection{Details of $\ourmodel$ performance w.r.t. Policy Network Sizes}
Table~\ref{app:tab:ab4} shows detailed experiment results of Table~\ref{tab:ab4}.
As the network size increases, overall performance generally increases. 
However, the network capacity of the reasoning-policy $\Phi_\text{R}$ has a more significant impact on enhancing performance compared to the planning-policy $\Phi_\text{P}$.

\begin{table*}[h]
\caption{Details of $\ourmodel$ performance w.r.t. policy network sizes}
\vskip 0.1in
\begin{center}
\begin{adjustbox}{width=0.95\textwidth}
    \begin{tabular}{ll cc cc cc cc cc}
    \toprule
    \multirow{2}{*}{$\Phi_\text{R}$} & \multirow{2}{*}{$\Phi_\text{P}$} & \multirow{2}{*}{Parameter Size}
    & \multicolumn{2}{c}{Train} & \multicolumn{2}{c}{Seen} & \multicolumn{2}{c}{Unseen Spatial} & \multicolumn{2}{c}{Unseen Environment} \\
    
    \cmidrule(rl){4-5} \cmidrule(rl){6-7} \cmidrule(rl){8-9} \cmidrule(rl){10-11}
     & & & SR  & GC  & SR  & GC  & SR  & GC  & SR  & GC  \\
    \midrule
    
    t5-small & gpt2
    & $0.06$B+$0.1$B
    & $100.0{\pm0.0}$ & $100.0{\pm0.0}$
    & $81.8{\pm0.5}$ & $92.2{\pm0.2}$
    & $52.7{\pm1.0}$ & $81.2{\pm0.4}$
    & $40.3{\pm0.9}$ & $68.7{\pm0.6}$ \\

    t5-small & gpt2-medium
    & $0.06$B+$0.4$B
    & $100.0{\pm0.0}$ & $100.0{\pm0.0}$
    & $81.4{\pm0.1}$ & $92.1{\pm0.1}$
    & $52.6{\pm0.4}$ & $81.0{\pm0.4}$
    & $39.7{\pm1.2}$ & $69.9{\pm0.0}$ \\

    t5-small & gpt2-large
    & $0.06$B+$0.8$B
    & $100.0{\pm0.0}$ & $100.0{\pm0.0}$
    & $81.8{\pm0.1}$ & $92.1{\pm0.1}$
    & $52.5{\pm0.2}$ & $81.2{\pm0.1}$
    & $67.5{\pm1.7}$ & $39.7{\pm2.0}$ \\

    t5-base & gpt2
    & $0.2$B+$0.1$B
    & $100.0{\pm0.0}$ & $100.0{\pm0.0}$
    & $79.4{\pm0.5}$ & $91.2{\pm0.3}$
    & $55.2{\pm0.7}$ & $82.8{\pm0.4}$
    & $41.8{\pm1.7}$ & $69.3{\pm1.2}$ \\

    t5-base & gpt2-medium
    & $0.2$B+$0.4$B
    & $100.0{\pm0.0}$ & $100.0{\pm0.0}$
    & $78.8{\pm0.6}$ & $91.0{\pm0.3}$
    & $55.2{\pm0.3}$ & $83.1{\pm0.3}$
    & $42.0{\pm1.0}$ & $69.6{\pm0.3}$ \\

    t5-base & gpt2-large
    & $0.2$B+$0.8$B
    & $100.0{\pm0.0}$ & $100.0{\pm0.0}$
    & $80.0{\pm1.3}$ & $91.5{\pm0.5}$
    & $54.9{\pm0.5}$ & $82.4{\pm0.6}$
    & $43.1{\pm2.4}$ & $70.2{\pm1.3}$ \\

    t5-large & gpt2
    & $0.7$B+$0.1$B
    & $100.0{\pm0.0}$ & $100.0{\pm0.0}$
    & $\mathbf{82.1{\pm1.0}}$ & $\mathbf{92.3{\pm0.3}}$
    & $57.2{\pm0.4}$ & $83.0{\pm0.1}$
    & $47.7{\pm1.0}$ & $73.5{\pm0.8}$ \\

    t5-large & gpt2-medium
    & $0.7$B+$0.4$B
    & $100.0{\pm0.0}$ & $100.0{\pm0.0}$
    & $81.8{\pm0.6}$ & $\mathbf{92.3{\pm0.2}}$
    & $\mathbf{57.5{\pm0.4}}$ & $\mathbf{83.2{\pm0.2}}$
    & $48.3{\pm1.4}$ & $73.5{\pm0.8}$ \\
    
    t5-large & gpt2-large
    & $0.7$B+$0.8$B
    & $100.0{\pm0.0}$ & $100.0{\pm0.0}$
    & $81.2{\pm0.4}$ & $92.0{\pm0.1}$
    & $\mathbf{57.5{\pm0.3}}$ & $83.1{\pm0.1}$
    & $\mathbf{49.1{\pm1.2}}$ & $\mathbf{74.2{\pm0.3}}$ \\
    
    \bottomrule
    \end{tabular}

\end{adjustbox}
\end{center}
\label{app:tab:ab4}
\vskip -0.1in
\end{table*}

\section{Embodied Knowledge Graph Examples}
\begin{figure*}[h]
\centering
   \includegraphics[width=0.4\linewidth]{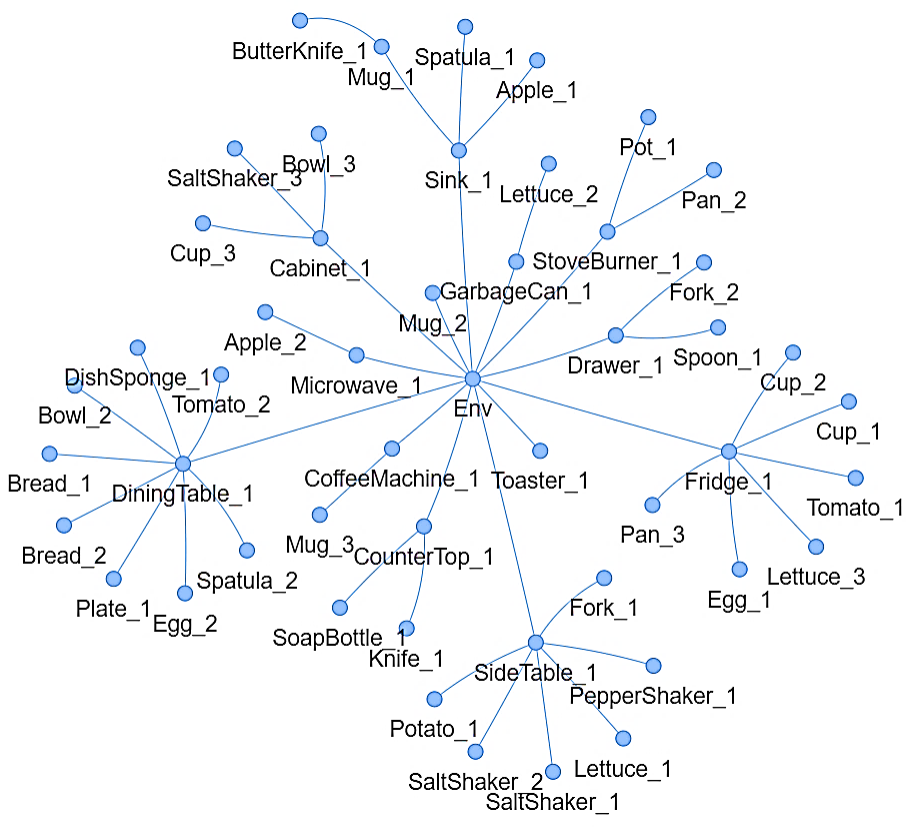}
   \vskip -0.05in
   \caption{Example of initial embodied kG}
   \label{fig:init_kg}
% \vskip -0.2in
\end{figure*}

\begin{figure*}[h]
\centering
   \includegraphics[width=0.90\linewidth]{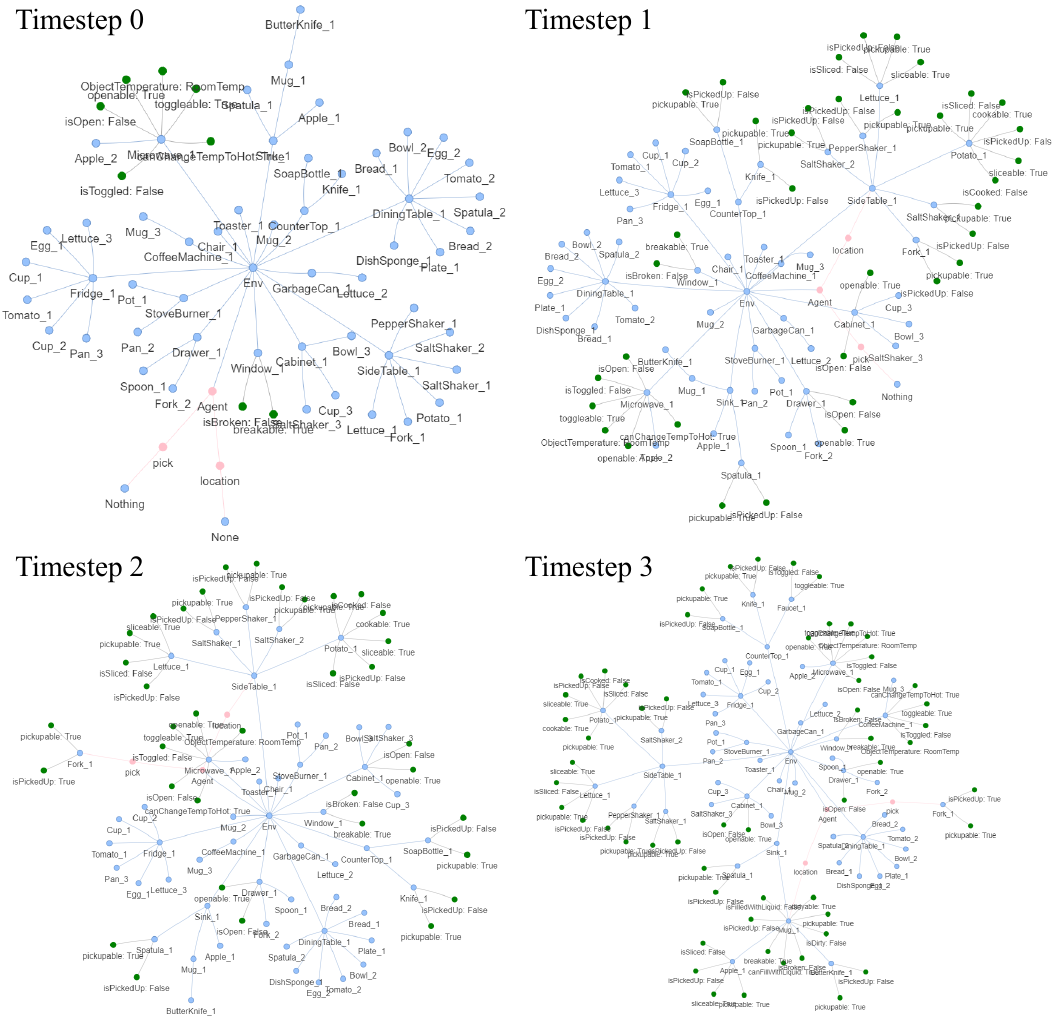}
   \vskip -0.05in
   \caption{Maintenance of the embodied KG to accommodate changes in environment information during task execution by the agent.}
   \label{fig:step_kg}
\vskip -0.2in
\end{figure*}

\begin{figure*}[h]
\centering
   \includegraphics[width=0.8\linewidth]{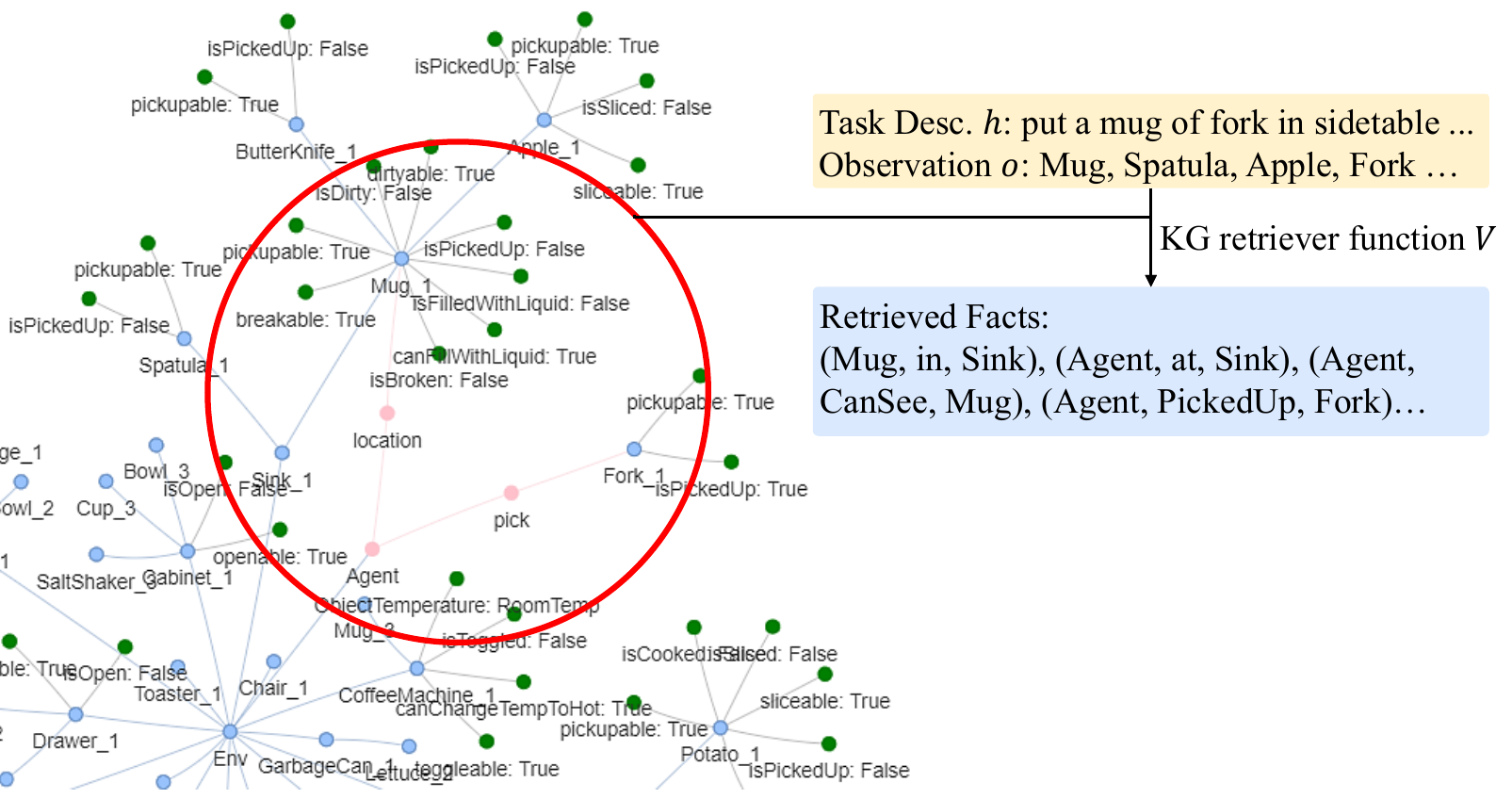}
   \vskip -0.05in
   \caption{Example of retrieved subset of embodied KG}
   \label{fig:init_retrieval}
\vskip -0.2in
\end{figure*}

Figures~\ref{fig:init_kg} and \ref{fig:step_kg} show examples of our embodied KG.
Initially, the task description provides information on a partially observable environment. 
From this information, our sLM-based policy efficiently encapsulates knowledge of the environment via an embodied KG as illustrated in Figure~\ref{fig:init_kg}. As the agent interacts with its surroundings, gathering information necessary for task completion, the embodied KG reflects those changes via the update function $U$, as specified in \eqref{eq:update}. Examples of these dynamic changes in the embodied KG, reflective of the evolving environment, are depicted in Figure\ref{fig:step_kg}. Finally, our KG retriever function $V$, detailed in \eqref{eq:kgretriever}, selects a subset of the embodied KG based on the task description $h$ and observation $o$. This subset is then used for embodied KG prompting to our reasoning-policy, as illustrated in Figure~\ref{fig:init_retrieval}.

\end{document}